%% file: insulin_resistance.tex
\documentclass[11pt, a4paper, logo, copyright]{googledeepmind}

\usepackage[authoryear, sort&compress, round]{natbib}
\input{math_commands.tex}

\usepackage{times}
\usepackage{latexsym}
\usepackage[T1]{fontenc}
\usepackage[utf8]{inputenc}
\usepackage{microtype}
\usepackage{inconsolata}
\usepackage{graphicx}
\usepackage{balance}       
\usepackage{graphics}      
\usepackage{hyperref}
\usepackage{color}
\usepackage{booktabs}
\usepackage{textcomp}
\usepackage{subcaption}
\usepackage{enumerate}
\usepackage{xcolor}
\usepackage{lipsum}
\usepackage{makecell}
\usepackage{multicol}
\usepackage{multirow}
\usepackage{array}
\usepackage{verbatimbox}
\usepackage{enumitem}
\usepackage{amsmath}
\usepackage{float}
\usepackage{stfloats}
\usepackage{graphicx}
\usepackage{amsthm}
\usepackage{listings}
\usepackage{caption} 
\usepackage[export]{adjustbox}
\usepackage{xspace}
\usepackage{epsfig}
\usepackage[linesnumbered]{algorithm2e}
\usepackage{algpseudocode}
\usepackage{tabularx}
\usepackage{arydshln}
\usepackage[bottom]{footmisc}
\usepackage{tcolorbox}
\usepackage{stackengine}
\usepackage{placeins}

\usepackage[normalem]{ulem}
\useunder{\uline}{\ul}{}
\tcbuselibrary{breakable}

\definecolor{E+F}{RGB}{	255, 99, 71}
\definecolor{B+F}{RGB}{255, 165, 0}
\definecolor{E+I}{RGB}{	173, 216, 230}
\definecolor{B+I}{RGB}{	30, 144, 255}
\definecolor{D}{RGB}{	60, 179, 113}

\definecolor{maroon}{cmyk}{0,0.87,0.68,0.32}

\usepackage{tikz}

\definecolor{darkgreen}{rgb}{0.0, 0.5, 0.0}
\definecolor{usercolor}{RGB}{200, 230, 250} 
\definecolor{coachcolor}{RGB}{180, 250, 180} 

\title{Insulin Resistance Prediction From Wearables and Routine Blood Biomarkers}
\reportnumber{} 
\renewcommand{\today}{2025-04-24}

\author[1,$\dagger$]{Ahmed A. Metwally}
\author[1]{A. Ali Heydari}
\author[1]{Daniel McDuff}
\author[1]{Alexandru Solot}
\author[1]{Zeinab Esmaeilpour}
\author[1]{Anthony Z. Faranesh}
\author[1]{Menglian Zhou}
\author[2]{David B. Savage}
\author[1]{Conor Heneghan}
\author[1]{Shwetak Patel}
\author[1]{Cathy Speed}
\author[1]{Javier L. Prieto}

\affil[1]{Google Research, Mountain View, CA, USA}
\affil[2]{Institute of Metabolic Science, University of Cambridge, Cambridge, UK}
\affil[$\dagger$]{Correspondence: aametwally@google.com}

\input{sections/0-abstract}
\begin{document}

\maketitle
\input{sections/1-introduction}
\input{sections/2-results}

\input{sections/3-discussion}

\pagebreak
\input{sections/4-methods}

\pagebreak
\section*{Acknowledgement}
We are deeply grateful to the Fitbit and Pixel Watch study participants who contributed their data to this research. We thank members of the Consumer Health Research Team at Google for their valuable feedback and technical support throughout this study, in particular Shelten Yuen, Anupam Pathak, Mark Malhotra, Jake Sunshine, Florence Thng, and Jiening Zhan. We thank Jacqueline Shreibati and Matthew Thompson from the Clinical team at Google for their feedback on the clinical utility and deployment of the proposed insulin resistance prediction model. We thank the software development team who built the service that was used to recruit this large cohort and enabled remote collection of wearables and blood biomarker data; Alex Dan, Alex Badescu, Delia-Georgiana Stuparu, George-Iulian Nitroi, Silviu Grigore, Paul Navin, and Dima Trubnikov. We also thank our collaborators at Quest Diagnostics. We thank Heiko Maiwand for helping in creating the graphical abstract of this manuscript. Special thanks to the project management team: Jen Galvan, Lourella Palao, and Sarah Wemmer. We also thank Brent Winslow, Nova Hammerquist, Amai Mai, Derek Peyton, and Edith Chung for their tremendous help setting up evaluation infrastructure for external endocrinologists to assess the insulin resistance agent. We thank Tracy Giest, Herschel Watkins, Lawrence Cai, Emily Blanchard, Ray Luo, Miao Liu, Jacob Gile, and the entire Human Research Laboratory at Google for their spectacular work in recruiting the validation cohort and collection of multimodality data.

\section*{Data and Code Availability}
Data and code used in this study will be made available upon the acceptance of this manuscript.

. 

\section*{Competing Interests}
This study was funded by Google LLC. A.A.M., A.A.H, D.M., A.S., Z.E., A.Z.F, M.Z., C.H., S.P., C.S., and J.L.P are employees of Alphabet and may own stock as part of the standard compensation package.

\pagebreak
\bibliography{insulin_resistance}

\renewcommand{\thesubsection}{Supplementary Table S\arabic{subsection}}
\setcounter{subsection}{0} 


\end{document}

%% file: math_commands.tex
\usepackage{amsmath,amsfonts,bm}

\def\eqref#1{equation~\ref{#1}}

\def\1{\bm{1}}

\DeclareMathAlphabet{\mathsfit}{\encodingdefault}{\sfdefault}{m}{sl}
\SetMathAlphabet{\mathsfit}{bold}{\encodingdefault}{\sfdefault}{bx}{n}



%% file: sections/0-abstract.tex
\begin{abstract}

Insulin resistance, a precursor to type 2 diabetes, is characterized by impaired insulin action in tissues. Current methods for measuring insulin resistance, while effective, are expensive, inaccessible, not widely available and hinder opportunities for early intervention. In this study, we remotely recruited the largest and most geographically, age and gender diverse dataset to date across the US to study insulin resistance (N=1,165 participants, with median BMI=28 kg/m$^2$, age=45 years, HbA1c=5.4\%), incorporating wearable device time series data and blood biomarkers, including the ground-truth measure of insulin resistance, homeostatic model assessment for insulin resistance (HOMA-IR). We developed deep neural network models to predict insulin resistance based on readily available digital and blood biomarkers. Our results show that our models can predict insulin resistance by combining both wearable data and readily available blood biomarkers better than either of the two data sources separately (R$^2$=0.5, auROC=0.80, Sensitivity=76\%, and specificity 84\%, based on HOMA-IR cut-off of 2.9, using wearables, demographics, and routine blood biomarkers). The model showed 93\% sensitivity and 95\% adjusted specificity in obese and sedentary participants, a subpopulation most vulnerable to developing type 2 diabetes and who could benefit most from early intervention. Adjusted specificity emphasizes false positives (predicting insulin resistance in truly insulin-sensitive individuals and not  impaired insulin sensitivity). Rigorous evaluation of model performance, including interpretability, robustness, and sensitivity, facilitates generalizability across larger cohorts, which is demonstrated by reproducing the prediction performance on an independent validation cohort (N=72 participants). Additionally, we demonstrated how the predicted insulin resistance can be integrated into a large language model agent to help understand and contextualize HOMA-IR values, facilitating interpretation and safe personalized recommendations. This work offers the potential for early detection of people at risk of type 2 diabetes and thereby facilitate earlier implementation of preventative strategies. Early identification could allow for lifestyle interventions that could reverse insulin resistance in those at highest risk of developing type 2 diabetes and could have a particularly profound impact among those who are unknowingly at risk. 

\end{abstract}

%% file: sections/1-introduction.tex
\section{Introduction}

Currently, 537 million adults are living with diabetes, a figure that is estimated to increase to 643 million by 2030, where approximately 10\% of people with diabetes have Type 1 diabetes (T1D), while around 90\% have Type 2 diabetes (T2D)\citep{1-noauthor_2022-vm}. This rise of T2D will be driven largely by lifestyle behaviors \citep{Galaviz2018-sa}. In a healthy individual, insulin, a hormone secreted by pancreatic beta-cell, helps regulate blood glucose levels by facilitating glucose uptake into cells (including muscles, adipose, and liver) from the blood. Additionally, incretin hormones, such as Glucagon-like Peptide-1 (GLP-1) and Gastric Inhibitory Polypeptide (GIP), can increase insulin secretion from pancreatic beta-cells, leading to improved glycemic control  \citep{El2021-dm}. The fundamental problem in diabetes is the inability of the body to regulate blood glucose properly due to absolute or relative insulin deficiency. In T1D, the body’s immune system mistakenly attacks and destroys the pancreatic beta-cells, resulting in absolute insulin deficiency and elevated blood glucose \citep{Roep2020-ne}. Alternatively, in the majority of T2D, the body becomes insulin resistant, requiring higher amounts of insulin to be produced by the pancreatic beta-cells to achieve the same glucose-lowering effect. With time, the pancreatic beta cells might become unable to produce enough insulin to compensate for insulin resistance (IR), leading to relative insulin deficiency and elevated blood glucose levels. The prevalence of IR in the general population is estimated to be between 20\% and 40\%, with variations observed across different ethnic groups, age brackets, lifestyle, and the presence of comorbidities \citep{Parcha2021-tg}. Chronic insulin resistance puts the person at significant risk for prediabetes and overt type 2 diabetes and is also strongly associated with metabolic dysfunction-associated steatotic liver disease (MASLD) and cardiovascular disease (CVD) risk. IR prevalence in type 2 diabetes is 83.9\% \citep{Bonora1998-og}. Contributing factors include excess body weight, particularly visceral fat, physical inactivity, and genetic predisposition. Figure \ref{fig:figure1}A illustrates the complex nature of T2D and the intricate relationships between lifestyle choices, genetics, and various metabolic subphenotypes and physiological processes involved in the development of the disease. Insulin resistance may manifest subtly with high blood pressure, or abnormal lipid profiles. Long term complications of diabetes include damage to various organs and tissues over time, such as diabetic retinopathy, nephropathy, and neuropathy \citep{Sapra2022-wn}. Moreover, insulin resistance is part of the pathophysiology of cardiovascular-kidney-metabolic (CKM) syndrome, characterized by interactions among metabolic risk factors, chronic kidney disease, and the cardiovascular system. Insulin resistance is a key component of metabolic syndrome, which contributes to the development of cardiovascular disease \citep{Khan2023-ye, Ndumele2023-as, Iglesies-Grau2023-mi}. Early detection of insulin resistance and intervention could play a critical role in reducing CKM disease burden. 

Identifying IR early can guide several focused lifestyle interventions, like weight loss, regular exercise, and a healthy dietary pattern, that can substantially improve or even reverse insulin resistance. While most individuals can certainly benefit from most types of physical activity and healthy diets, there are specific interventions that have been scientifically proven to prevent and treat IR. For lifestyle interventions, aerobic training \citep{Turcotte2008-qr}, resistance training \citep{Turcotte2008-qr, Niemann2020-lz}, calorie-restricted diets \citep{Zhang2021-wn}, and low-fat diets \citep{Demaria2023-in} have all shown to be valuable in reducing IR. On the therapeutic end, Thiazolidinediones and Metformin medications have frequently been shown to reduce IR \citep{Ko2017-la, Sui2019-ys}. Recent studies have shown that incretin hormone agonists, like GLP-1 and GIP, act as sensitizers and improve insulin resistance \cite{Garvey2022-qd, Frias2024-ct}.

Several methods are available for measuring insulin resistance (IR), but are not implemented routinely, meaning that opportunities for early intervention are often missed. Instead, a focus on snapshots of glucose levels, fasting glucose,  Hemoglobin A1c (HbA1c), or glucose level after a 2-hour oral glucose tolerance test (OGTT) represent the typical screening approach, and can be insensitive to those in early stages of IR. The gold standard test for IR is the hyperinsulinemic euglycemic clamp \citep{DeFronzo1979-xk}, which is performed in research facilities only, expensive, and time-consuming. Homeostatic Model Assessment of Insulin Resistance (HOMA-IR) and minimal model-based glucose and insulin measures are affordable and faster alternatives that can be performed in clinical labs \citep{Matthews1985-wg}. However, they require a clinical lab visit and may not be as accurate as the clamp. HOMA-IR is a mathematical model used to estimate the degree of insulin resistance in an individual \citep{Matthews1985-wg}. It provides a simple way to assess how well the body is using insulin. There is no single defined cut-off value, but a HOMA-IR above 2.9 is often suggestive of insulin resistance, HOMA-IR below 1.5 is suggestive of insulin sensitivity, and HOMA-IR between 1.5 and 2.9 is suggestive of impaired insulin sensitivity (early signs of insulin resistance)\citep{Parcha2021-tg, Gayoso-Diz2013-pf, Endukuru2020-lj, Da_Silva2023-tm}. Glucotyping, which is a framework that analyzes glucose-time series trends and clusters them into patterns called glucotypes based on glucose variability, via continuous glucose monitoring (CGM) sensors is a promising new method to detect insulin resistance that can be done at home, but it requires further validation studies \citep{Hall2018-nn, Metwally2024-cgm}. Physiological signals derived from smartwatches, including heart rate derived metrics (e.g. resting heart rate “RHR”, heart rate variability “HRV”), physical activity, and sleep metrics may conceivably help to predict IR, since they have been found to affect glucose regulation \citep{Eckstein2022-gx, Griggs2022-zk}. It has been shown that higher RHR and lower HRV are associated with IR \citep{Saito2015-ib, Svensson2016-ts, Poon2020-cj, Flanagan1999-sa, Beddhu2009-yw, Saito2022-fk, Grandinetti2015-ql}. 

In this work, we present a method for predicting insulin resistance using (a) signals derived from a consumer smartwatch (e.g., resting heart rate, heart rate variability, step count, sleep duration), (b) demographics (e.g., BMI, age), and (c) commonly measured blood biomarkers such as lipid panels (e.g. LDL, triglycerides, and HDL). This method has the potential to be scaled to millions of individuals, enabling widespread identification of IR. We assembled the largest cohort to date ($N=1,165$) with a combined wearable lifestyle, demographics, blood biomarker dataset, and a ground truth IR measure (HOMA-IR). A comprehensive analysis of our machine learning (ML) model was conducted, including interoperability, stratification, and robustness analyses, to quantify its generalizability and scalability. Furthermore, we developed a large language model (LLM) agent that utilizes the IR model's output, along with participant lifestyle and blood biomarker data, to provide safe, holistic insights into individual metabolic health and diabetes risk, and offers personalized recommendations and illustrative explanations.

%% file: sections/2-results.tex
\section{Results}
\label{sec:results}

\subsection{Study design and cohort characteristics}
We designed a prospective observational study and recruited adult participants from the United States to participate in a consented research study. The study was approved by Advarra as the institutional review board (IRB) of record (protocol number Pro00074093) and conducted remotely using the Google Health Studies (GHS) application. GHS is a consumer-facing secure platform for conducting digital studies that allows participants to enroll, check eligibility, and provide informed consent (Methods). Additionally, in this instance it was configured to enable the collection of wearable data from Fitbit and Google Pixel Watch devices (collectively referred throughout as “wearables”), completion of questionnaires, and ordering of blood tests with Quest Diagnostics. Participants were asked to: (1) link their Fitbit/Google account to the GHS app and share data, (2) complete questionnaires about demographics, health history, and health information, and (3) schedule and complete a blood draw at a Quest Patient Service Center within 65 days of enrollment.

\begin{table}[]
\centering
\caption{\textbf{Participant Characteristics.} Continuous variables are presented as median (standard deviation) and categorical variables as counts.}
\label{tab:table_1}
\resizebox{\textwidth}{!}{%
\begin{tabular}{ccccccc}
\hline
\multicolumn{1}{l}{}                       & \multicolumn{1}{l}{}     & \textbf{ALL}    & \textbf{IS}     & \textbf{IMPAIRED-IS} & \textbf{IR}     & \textbf{P-Value}          \\ \hline
\textbf{\# of Participants}                & N                        & 1165            & 459             & 406                  & 300             & \multicolumn{1}{l}{}      \\ \hline
\multirow{2}{*}{\textbf{Demographics}}     & AGE                      & 45.0 (12.5)     & 45.5 (12.0)     & 44.0 (12.5)          & 47.5 (12.9)     & 0.11                      \\ \cline{2-7} 
                                           & BMI                      & 28.0 (6.7)      & 24.4 (4.3)      & 27.8 (5.7)           & 32.6 (7.4)      & 1.95E-64                  \\ \hline
\multirow{4}{*}{\textbf{Gender}}           & Female                   & 636             & 146             & 331                  & 159             & \multirow{4}{*}{0.52}     \\ \cline{2-6}
                                           & Male                     & 505             & 119             & 251                  & 135             &                           \\ \cline{2-6}
                                           & Others                   & 21              & 7               & 8                    & 6               &                           \\ \cline{2-6}
                                           & Choose not to answer     & 3               & 0               & 3                    & 0               &                           \\ \hline
\multirow{8}{*}{\textbf{Ethnicity}}        & White / Caucasian        & 905             & 222             & 460                  & 223             & \multirow{8}{*}{0.18}     \\ \cline{2-6}
                                           & Hispanic                 & 67              & 9               & 38                   & 20              &                           \\ \cline{2-6}
                                           & Asian - Indian           & 54              & 4               & 34                   & 16              &                           \\ \cline{2-6}
                                           & Asian - Eastern          & 31              & 9               & 14                   & 8               &                           \\ \cline{2-6}
                                           & African-American         & 46              & 15              & 20                   & 11              &                           \\ \cline{2-6}
                                           & Native-American          & 4               & 1               & 2                    & 1               &                           \\ \cline{2-6}
                                           & Mixed Race               & 38              & 8               & 16                   & 14              &                           \\ \cline{2-6}
                                           & Choose not to answer     & 20              & 4               & 9                    & 7               &                           \\ \hline
\multirow{4}{*}{\textbf{Wearables}}        & RHR (bpm during sleep)   & 66.0 (8.2)      & 62.0 (7.1)      & 67.0 (7.9)           & 70.0 (8.0)      & 1.19E-21                  \\ \cline{2-7} 
                                           & SLEEP DURATION (minutes) & 459.0 (66.0)    & 460.0 (52.2)    & 464.0 (64.3)         & 444.8 (77.3)    & 1.33E-05                  \\ \cline{2-7} 
                                           & STEPS (Daily)            & 6909.0 (3752.6) & 8395.0 (4491.9) & 6974.5 (3383.1)      & 5596.5 (3179.3) & 4.37E-18                  \\ \cline{2-7} 
                                           & HRV (RMSSD during sleep) & 27.1 (16.5)     & 29.6 (19.7)     & 27.6 (15.4)          & 24.3 (14.8)     & 2.03E-06                  \\ \hline
\multirow{5}{*}{\textbf{Blood Biomarkers}} & HBA1C                    & 5.4 (0.5)       & 5.3 (0.3)       & 5.4 (0.4)            & 5.6 (0.7)       & 4.95E-38                  \\ \cline{2-7} 
                                           & GLUCOSE                  & 90.0 (13.2)     & 84.0 (7.3)      & 90.0 (9.3)           & 98.0 (17.4)     & 4.55E-66                  \\ \cline{2-7} 
                                           & HDL                      & 56.0 (15.4)     & 62.0 (17.1)     & 57.0 (14.2)          & 49.0 (12.4)     & 7.28E-33                  \\ \cline{2-7} 
                                           & LDL                      & 105.0 (34.2)    & 102.0 (31.1)    & 109.0 (34.3)         & 104.0 (36.1)    & 3.11E-05                  \\ \cline{2-7} 
                                           & TRIGLYCERIDES            & 89.0 (61.8)     & 65.0 (32.4)     & 89.0 (53.4)          & 124.5 (76.6)    & 6.01E-48                  \\ \hline
\multirow{6}{*}{\textbf{Comorbidities}}    & CVD                      & 37              & 2               & 21                   & 14              & \multirow{6}{*}{4.88E-07} \\ \cline{2-6}
                                           & HYPERLIPIDEMIA           & 247             & 43              & 125                  & 79              &                           \\ \cline{2-6}
                                           & DIABETES                 & 64              & 2               & 16                   & 46              &                           \\ \cline{2-6}
                                           & RESPIRATORY              & 154             & 29              & 67                   & 58              &                           \\ \cline{2-6}
                                           & HYPERTENSION             & 246             & 32              & 101                  & 113             &                           \\ \cline{2-6}
                                           & KIDNEY DISEASE           & 23              & 5               & 12                   & 6               &                           \\ \hline
\end{tabular}
}
\end{table}

\begin{figure*}
    \centering
    \includegraphics[width=0.88\textwidth]{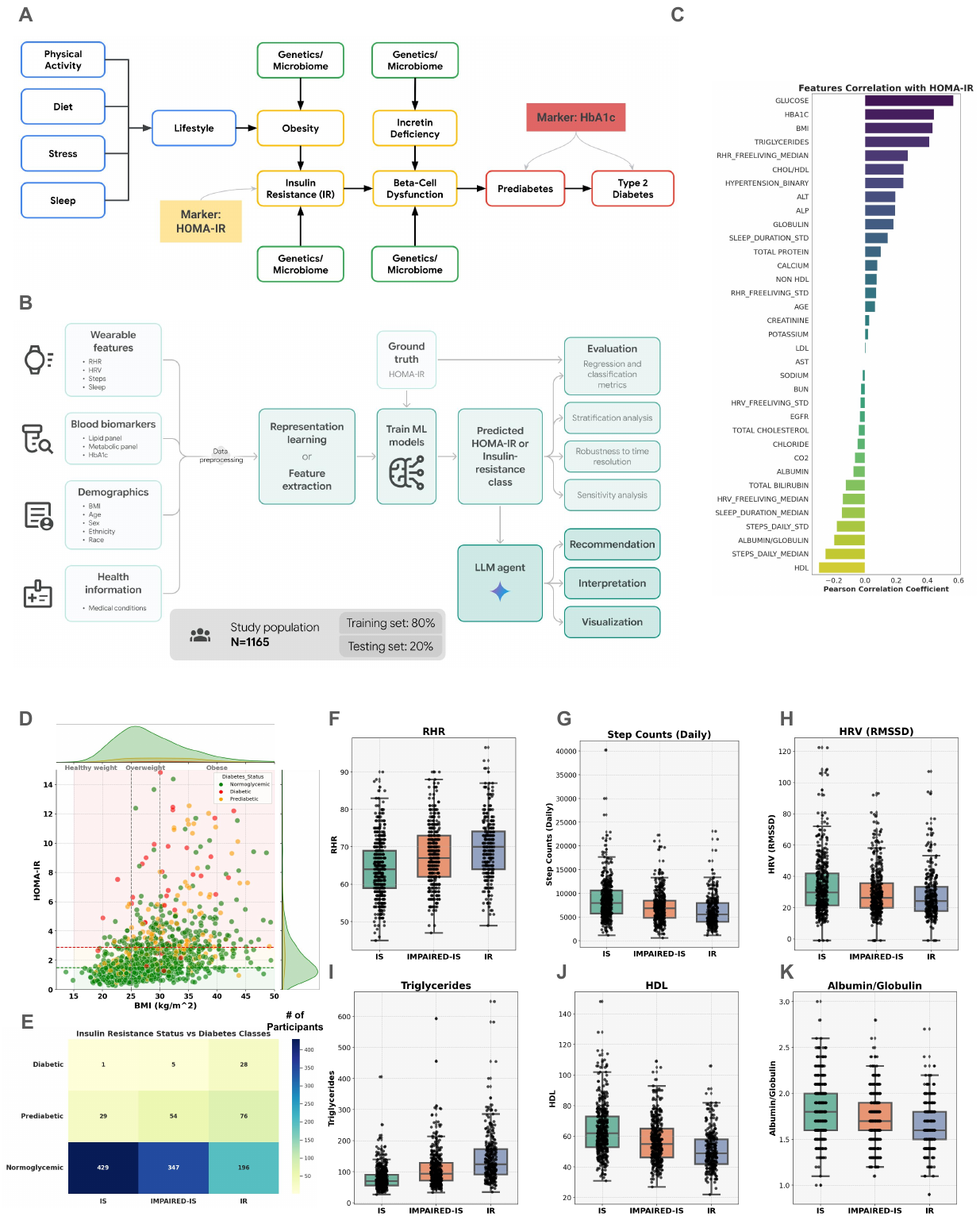}
    \begin{minipage}{\textwidth}
        \caption{\textbf{Study Design and Data Summary.} (A) Overview of physiological factors and associated lifestyle factors leading to insulin resistance, prediabetes and diabetes. (B) Illustration of our proposed modeling pipeline for predicting HOMA-IR, and interpreting the results with the Insulin Resistance Education and Understanding Large Language Model. (C) Correlation of blood biomarkers and lifestyle features (continuous values) with HOMA-IR. (D) Scatter plot of BMI and HOMA-IR values which signifies the relationship between higher BMI values and insulin resistance (measured through HOMA-IR). (E, F, G) Distribution of top three highly-correlated wearables features with HOMA-IR, namely Resting Heart Rate, Daily Step Counts, and Heart Rate Variability, for stratified insulin sensitivity groups (insulin sensitive, impaired insulin sensitivity, and insulin resistance). (H, I, J) Distribution of top three highly-correlated blood biomarkers (Triglycerides, HDL and LDL cholesterol) for stratified insulin sensitivity groups (same as E, F, G).}
        \label{fig:figure1}
    \end{minipage}
\end{figure*}

We enrolled 4,416 participants, of which 1,165 (25\% completion rate) had complete data (at least 14 days of wearable data, blood biomarkers, and demographics, medical conditions), and passed preprocessing, and were included in our analysis (Methods, Supplementary Figure S1 and S2). Inclusion criteria included U.S. residents between the ages of 21 and 80 who wear a Fitbit wearable device or a Pixel Watch with heart rate sensing capabilities and are willing to go to a Quest Diagnostics location for blood draws. Participants were asked to do blood draws early in the morning after fasting for at least 8 hours to minimize the effect of the solar diurnal cycle. All collected data are listed in the Methods.

We used HOMA-IR, calculated as (HOMA-IR = [Fasting Insulin (µU/ml) × Fasting Glucose (mg/dL) ] / 405), as the ground truth for quantifying insulin resistance (Methods). Thresholds of HOMA-IR used to define insulin resistance vary widely in the literature, ranging from 2.5–3.5 for significant insulin resistance to 1–1.5 for insulin sensitivity \citep{Parcha2021-tg, Gayoso-Diz2013-pf, Endukuru2020-lj, Da_Silva2023-tm}. The differences in thresholds are primarily attributed to differences in study populations (ethnicity, age, and gender), and the specific factor used to define the threshold (either maximizing the sensitivity for predicting metabolic syndrome or using the 90th percentile). In this study, participants were classified as insulin resistant (IR) (HOMA-IR > 2.9), insulin sensitive (IS) (HOMA-IR < 1.5), or impaired insulin sensitivity (Impaired-IS) (1.5 $ \leq $ HOMA-IR $ \leq $ 2.9). Table \ref{tab:table_1} summarizes our cohort characteristics. A total of 1,165 individuals (459 IS, 406 Impaired-IS, 300 IR) with high-quality data (Supplementary Figure S1) were included in the insulin resistance model development and validation. Compared to IS participants, those with IR had significantly higher BMI (pvalue <0.001, Supplementary Figure S3), RHR (pvalue <0.001), fasting glucose (pvalue <0.001), and triglycerides (pvalue <0.001), and lower step counts (pvalue <0.001). They also had higher rates of diabetes, CVD, hyperlipidemia, and hypertension. Supplementary Table S1 summarizes all digital and blood biomarkers between the three groups.

Figure \ref{fig:figure1}B illustrates our design of the deep learning framework for predicting insulin resistance. It takes various combinations of wearable data, blood biomarkers, demographics, and health information as input. Time-series data are preprocessed, summarized, and then an embedded representation is extracted via a masked autoencoder (Methods). This representation is fed into multiple tree-based models to predict continuous HOMA-IR values. For comparison and additional interpretability of the features, we trained a direct regression-based model to predict the HOMA-IR values as well. Insulin resistance classes (IS, Impaired-IS, IR) were obtained by thresholding on the predicted HOMA-IR using two thresholds 1.5 and 2.9. We tested 25 various combinations of input features (wearable features, demographics, fasting glucose, lipid panel, HbA1c, metabolic panel, and hypertension status). Training and testing the models were done using 5-fold cross-validation, as well as leave-one-out cross validation for direct regression models. Interpretability analysis was performed to assess feature contributions to the learned representation and the model performance. We performed a comprehensive evaluation of the predicted HOMA-IR and insulin resistance classes to assess the generalizability and scalability of the trained model. Lastly, we performed a robustness and stability analysis to assess the variability of the model's output for each individual in the dataset, based on varying time intervals of wearable data (1 week, 2 weeks, and up to 3 months). This analysis also highlights the optimal time interval of wearable data that can be used to predict insulin resistance. To showcase the potential of our approach for patient education and understanding, we developed a novel LLM-based metabolic health agent incorporating the trained HOMA-IR model. This agent provides personalized interpretations, recommendations, and visualizations, and can primarily be used to answer queries about diabetes and metabolic health.

\subsection{Insulin resistance association with lifestyle factors, blood biomarkers, and diabetes}

\begin{figure*}
    \centering
    \includegraphics[width=0.95\textwidth]{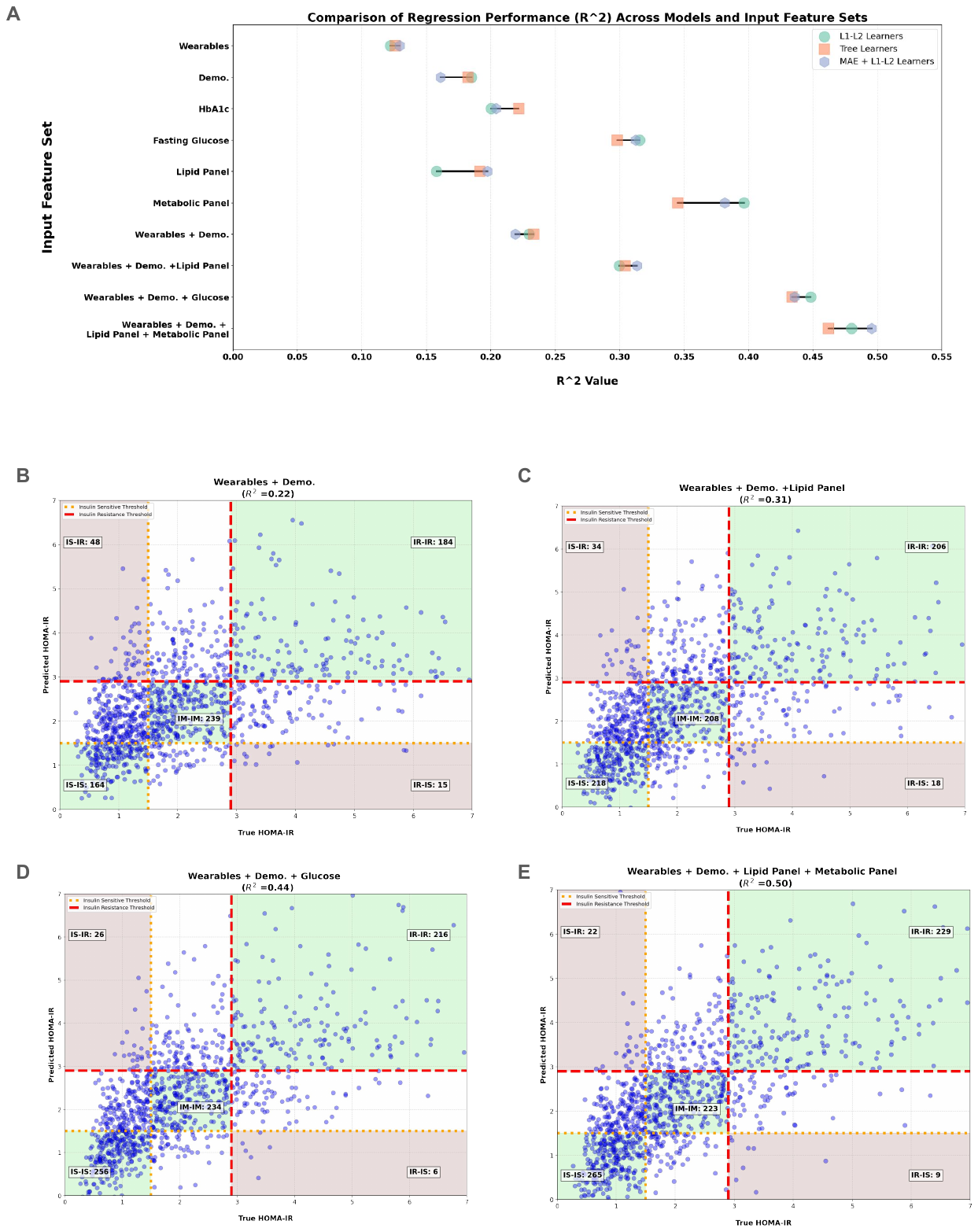}
    \caption{\textbf{Performance Evaluation of HOMA-IR prediction (Regression).} (A) Comparison of HOMA-IR regression across input feature sets and models. (B, C, D, E) Scatter plots of predicted HOMA-IR values versus the true HOMA-IR models for selected feature sets. Areas of concern for true positive and false negative are highlighted as light green and rosy brown, respectively.}
    \label{fig:figure2}
\end{figure*}

We calculated Pearson correlation coefficients between HOMA-IR and major lifestyle factors (RHR, HRV, step count, sleep duration), demographics, lipids, glucose, kidney and liver function markers, and key electrolytes. Figure \ref{fig:figure1}C demonstrates significant positive correlations between HOMA-IR and glucose (r=0.57, pvalue<0.001), BMI (r=0.43, pvalue<0.001), HbA1c (r=0.45, pvalue=9x10-53), triglycerides (r=0.40, pvalue=<0.001, Figure \ref{fig:figure1}I), and RHR (r=0.27, pvalue=<0.001, Figure \ref{fig:figure1}F). Moreover, HOMA-IR is significantly and negatively correlated with HDL cholesterol (r=-0.30, pvalue=<0.001, Figure \ref{fig:figure1}J), daily step count (r=-0.25, pvalue=<0.001, Figure \ref{fig:figure1}G), albumin/globulin ratio (r=-0.18, pvalue=<0.001, Figure 1K), and HRV (r=-0.14, pvalue<0.001, Figure \ref{fig:figure1}H). This suggests HOMA-IR could be inferred utilizing readily available measures from wearables or electronic health records (EHR). Age, kidney markers (e.g., creatinine, eGFR, blood urea nitrogen “BUN”), and electrolytes (e.g., Sodium, Potassium, and Chloride) have small Pearson correlation coefficient with HOMA-IR (|r|<0.1) (Supplementary Table S2). C-reactive protein (CRP) levels were significantly elevated in the insulin-resistant (IR) group compared to the insulin-sensitive (IS) group (2.8 mg/dL vs. 0.6 mg/dL, pvalue < 0.001). Analytes assessed in the standard complete blood count (CBC) (e.g., white blood cell, red blood cell, hemoglobin, hematocrit, etc), did not differ significantly in their effect size between the IR and IS groups (Supplementary Table S1). Figure \ref{fig:figure1}D illustrates the relationship between obesity (measured by BMI) and insulin resistance (IR), as assessed by HOMA-IR. 205/458 (45\%) of obese individuals (BMI>30) are insulin resistant (HOMA-IR>2.9). Only 22/319 (6.9\%) of participants with normal healthy weight (18.5<BMI<25) are insulin resistant. Figure \ref{fig:figure1}E highlights the relationship between IR and diabetes. In our cohort, 33/34 (97\%) of individuals with diabetes (HbA1c>6.5\%) are either insulin resistant or have impaired insulin sensitivity, while only 1 participant is insulin sensitive. Notably, many individuals may have IR without apparent elevated HbA1c levels at the time of the study. 196/972 (20\%) of normoglycemic participants are IR, representing those at high risk of developing diabetes. This underscores the importance of identifying these individuals early for personalized lifestyle interventions that could potentially reverse the course of T2D development. Supplementary Figure S4 shows pairwise correlation between HOMA-IR, wearable features, demographics, and blood biomarkers.

\begin{figure*}
    \centering
    \includegraphics[width=0.90\textwidth]{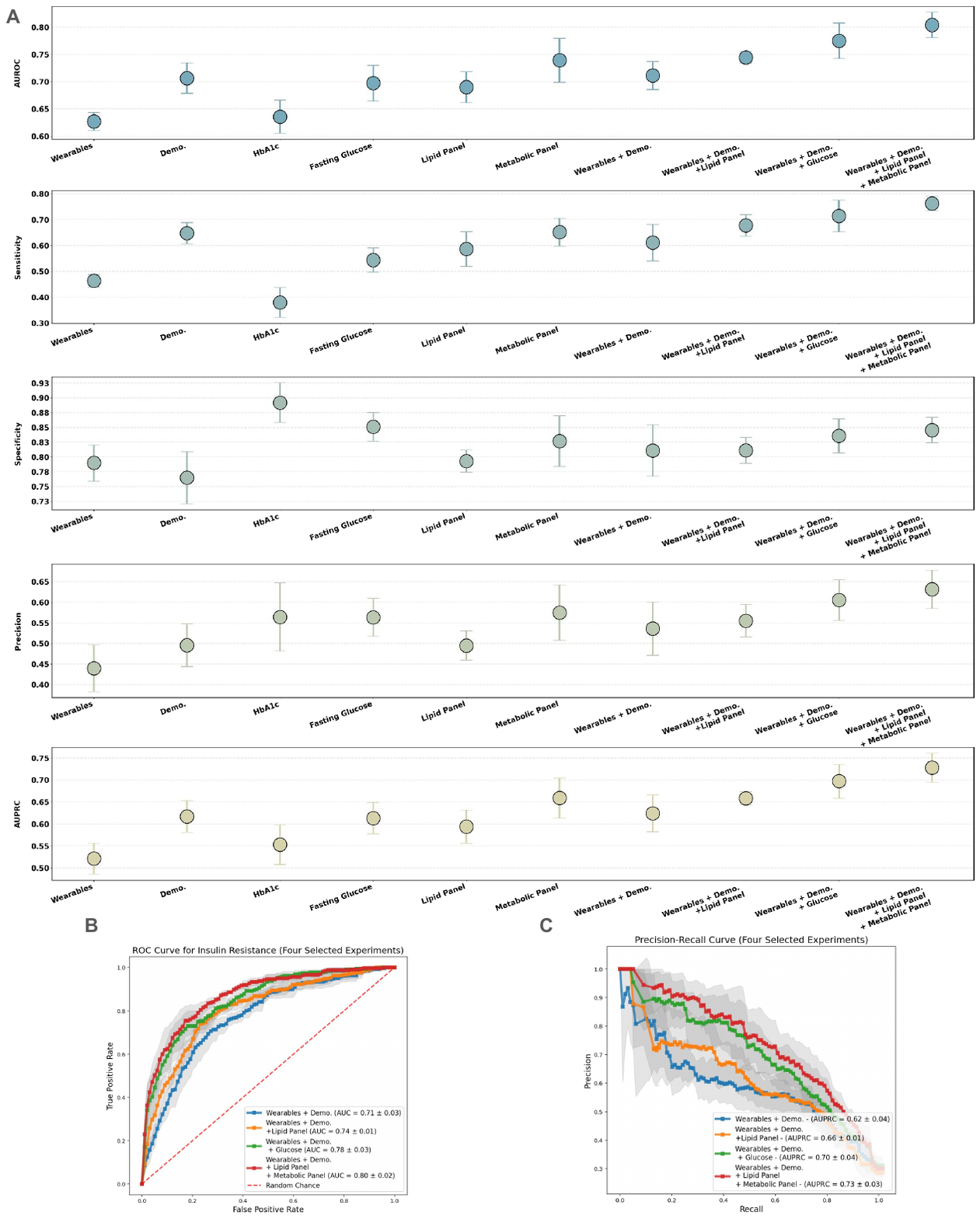}
    \caption{\textbf{Performance Evaluation of Insulin Resistance Prediction (Classification).} (A) Performance of our binary classification model for various input features for identifying insulin resistant individuals (using MAE + L1-L2 learners), measured through Area Under the Receiver Operating Characteristic curve (AUROC), Sensitivity, Specificity, Precision, and Area Under the Precision-Recall Curve (AUPRC). (B) Visualization of the ROC curves for various feature sets across five cross validation folds. Average values are colors, with the gray areas around each line indicating the standard deviation across the five folds. (C) Visualization of the Precision-Recall curve for selected feature sets. Average values are colors, with the gray areas around each line indicate the standard deviation across the five folds.}
    \label{fig:figure3}
\end{figure*}

\subsection{Insulin resistance prediction from wearables and blood biomarkers}
To identify the impact of different features and underlying models on predicting insulin resistance, we performed a large-scale ablation study, encompassing 25 different feature sets (“experiments”) (Supplementary Table S3), six different time windows, and four learning schemes, resulting in 600 unique model combinations. We assessed both traditional machine learning (ML) and deep learning (DL) approaches, utilizing 5-fold and leave-one-out cross-validation (CV) for ML models and 5-fold CV for DL models (Methods). For each model, we predicted continuous HOMA-IR values and applied classification thresholds to determine insulin resistance. We evaluated model performance using both regression metrics (R$^2$, mean absolute error “MAE”, mean squared error “MSE”) and classification metrics (Specificity, Sensitivity, Precision, AUROC, AUPRC) averaged across test set predictions from all cross-validation folds (Methods). 

Our initial analysis focused on determining the optimal CV strategy for our ML-based models . In pairwise comparisons across all classification metrics, 5-fold CV consistently outperformed leave-one-out CV, exhibiting both higher average performance and significantly lower variance (p-value < 0.05 for “Wearables + Demo”, “Wearables + Demo + Lipid Panels”, “Wearables + Demo + Glucose”, and “Wearables + Demo + Lipid Panel + Metabolic Panel”), where insulin was excluded from the features used in any of the tested models throughout this study. Specifically, 5-fold CV yielded higher average specificity, sensitivity, and precision in all unique experimental setups. Additionally, it demonstrated superior AUPRC in 193 instances (~60\%) and AUROC in 263 experiments (~81\%). While 5-fold CV performed slightly worse than leave-one-out CV in terms of regression metrics achieving slightly lower R$^2$ and higher MAE in approximately 46\% of the experiments, its lower variance across the same experiments led us to select it as the baseline for all subsequent analyses. Models incorporating a combination of wearable features, demographics, and fasting glucose consistently outperformed other models (shown in Figure \ref{fig:figure2}). We present a subset of our results in Figure \ref{fig:figure2}A (Supplementary Table S4), with the complete results of our ablation study available in the Supplementary Table S5 and S6.

We further investigated the effect of various window sizes (days preceding the lab test = {7, 14, 30, 60, 90}) on wearable feature aggregation (Methods). Using R$^2$ as the metric of interest, our results did not reveal an "optimal" number of days to consider, with performance plateauing after 30 days. Notably, using seven and 14 days consistently yielded competitive results across different feature sets. Given our results for accuracy and consistency (discussed later), and the importance of early intervention, we chose to use seven days as the default aggregation window. 

To further enhance predictive performance, we explored DL-based approaches for learning representations of the feature sets prior to regression (Methods). As shown in Figure \ref{fig:figure2}A, performing regression on these deep-learned representations (Masked Autoencoder “MAE” + L1-L2 Learner), as opposed to directly using raw feature sets, yielded improvements in some experiments. Specifically, in 20 of our 25 experiments, we saw improvements in R$^2$ and average MAE across cross validation folds against either direct regression methods(see Supplementary Table S4 for complete performance metrics). These improvements were consistent across both regression and classification metrics, underscoring the ability of representation learning to capture complex relationships within the data. Based on these findings, we selected two primary models for our remaining analyses: XGBoost with tree learners for direct regression and Masked Autoencoder + XGBoost with linear learners for representation learning.

Figure \ref{fig:figure2} illustrates the regression performance for selected feature sets using a seven-day aggregation window for wearables features. Figure \ref{fig:figure2}A displays R$^2$ for both direct regression (XGBoost with linear “L1-L2” and non-linear “Tree” learners) and representation learning (Masked Autoencoder “MAE” + XGBoost with linear learners “L1-L2”). Our results demonstrate that incorporating readily available blood biomarkers alongside wearables features and demographics significantly enhances prediction accuracy. Furthermore, our analyses show an increase of true positives (shaded green) and the reduction of consequential false predictions (shaded light brown) of Figure \ref{fig:figure2}B-E. Most notably, the addition of fasting glucose alone doubled the R$^2$ value (from 0.212 to 0.435, Figure \ref{fig:figure2}D), increased the number of correctly identified individuals with insulin resistance by 24\% (from 184 to 229), and minimized consequential false positives (participants who are identified as IR, but they are IS 54\% (from 48 to 22). Meanwhile, our experiments showed that using glucose alone is not sufficient (R$^2$ of 0.31, highlighting the importance of other lifestyle factors in estimating HOMA-IR as well. The optimal model for predicting HOMA-IR is when we combine wearable features, demographics, and readily available blood biomarkers (fasting glucose, lipid panel, metabolic panel) (R$^2$ of 0.50). The number of consequential false positives reduced to 22 positives (Figure \ref{fig:figure2}E).

Subsequently, we performed a rigorous evaluation of the models' capacity to accurately classify insulin resistance, employing the predicted HOMA-IR values with a threshold of 2.9. Figure \ref{fig:figure3}A shows the model based on wearables and demographics data alone can predict IR with auROC=0.70, sensitivity=0.60, and specificity=0.80. The incorporation of fasting glucose levels into this model resulted in a marked improvement in performance, yielding a significant increase in the performance (auROC=0.78, sensitivity=0.73, specificity=0.84). The optimal model which includes wearables, demographics, readily available blood biomarkers (lipid panel, and, metabolic panel) has auROC=0.80, sensitivity=0.76, and specificity=0.84. It is crucial to underscore that relying solely on demographics, wearables, fasting glucose, or lipid panels in isolation does not yield adequate predictive power for insulin resistance (Figure \ref{fig:figure3}A). Supplementary Table S7 reports the statistical significance of the differences in AUROC between each pair of experiments, as determined by McNemar's test. Supplementary Table S8 reports the p-values based on the Wilcoxon rank sum test. To further elucidate the performance of the models under different predicted HOMA-IR thresholds, Figures \ref{fig:figure3}B and \ref{fig:figure3}C present the ROC and Precision-Recall curves, respectively, for the four feature sets that can be practically implemented based on available data. Our findings demonstrate that the integration of wearable data, demographics, and readily accessible blood biomarkers significantly enhances our ability to predict insulin resistance compared to relying on each data source in isolation. 

To better quantify the generalizability of our approach on different ethnicities, we performed three additional experiments (using only the non-linear XGboost model) to measure the generalizability of our models to ethnicities included in our initial study cohort (Supplementary Table S9, Methods). Our stratified analysis by ethnicity reveals that our models, which predominantly trained on Caucasian subpopulation, can generalize well (on par with baseline models trained and tested on the entire population) to the Hispanic, Asian-Eastern and “mixed” ethnicities. It also shows that having diverse representation in the training set can impact the performance on other subpopulations, particularly African-American and Asian-Indian. When individuals from African American and Asian-Indian are included in the 80\% training split (Experiment 1), the testing performance increases, such as African-American (sp=0.91, se=1.0). We believe that this disparity underscores the importance of including even small numbers of individuals from these ethnicities during training. In Experiment 3, where models were trained on all ethnicities except the target ethnicity, our results showed improvements over Experiment 2 where we only trained on Caucasian individuals, showing that including other ethnicities in training can improve generalizability to unseen demographics (e.g. African American). However, this experiment further confirmed that including these ethnicities in the initial training data (Experiment 1) would improve performance of our models for some ethnicities.

\subsection{Interpretability of the learned representation and prediction model
}
\begin{figure*}
    \centering
    \includegraphics[width=0.90\textwidth]{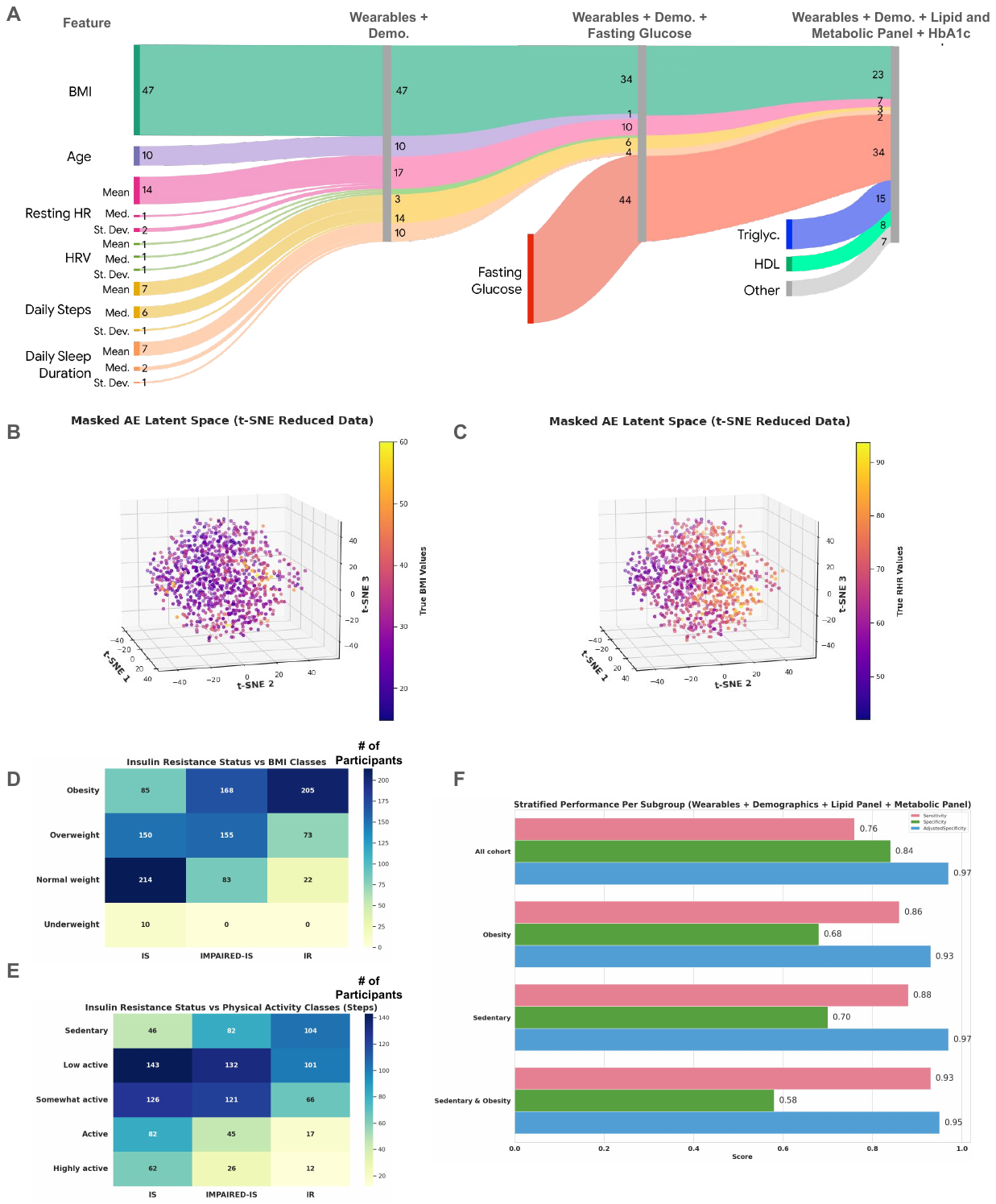}
    \caption{\textbf{Interpretability and Stratification.}  (A) Sankey diagram showing the relative feature importance (SHapley Additive exPlanations [SHAP] values) for each of the proposed nonlinear XGBoost models for direct regression. (B, C) Qualitative evaluation of learned latent space’s interpretability of learning important features. The t-SNE reduced latent space shows that individuals with higher BMI and resting heart rate are clustered closely together in space, following our quantitative results of classifying high BMI and high-RHR individuals using these learned representations. (D, E) Distribution of individuals stratified by insulin resistance class and BMI classes (D) and insulin resistance versus physical activity classes as determined by number of daily steps (E). (F) Results of classification performance for various lifestyle stratification.}
    \label{fig:figure4}
\end{figure*}

To gain insights into the models' decision-making processes and validate their physiological relevance, we performed feature importance analysis for both our direct regression (XGBoost) (Figure \ref{fig:figure4}A) and representation learning models (Figure \ref{fig:figure4}B and \ref{fig:figure4}C). For the XGBoost models, we leveraged Shapley Additive Explanations (SHAP) (Methods) \citep{Lundberg2017-ls} to quantify feature contributions across all models. In models incorporating either the full metabolic panel or fasting glucose alone, fasting glucose emerged as the most contributing feature, with BMI consistently ranking as the second most important feature in these models. Notably, in models without fasting glucose, BMI remained the dominant contributor to model predictions, highlighting its robust association with insulin resistance. When wearables-derived digital markers were included, resting heart rate consistently emerged within the top three most important features, ranking second after BMI when glucose was not included. Other wearables features, such as the median of daily steps and average sleep duration, also ranked among the top five most contributing factors. We provide a select panel of feature importance in Figure \ref{fig:figure4}A. Our findings align with both our empirical correlation analysis between HOMA-IR and wearables features, demographics, and blood biomarkers (Figure \ref{fig:figure1}C).

Interpreting the learned representations from our deep learning models required a different approach due to the indirect relationship between input features and model predictions. We employed both a qualitative and quantitative analysis of the learned latent space in order to validate the importance of features found in our direct regression experiments (Methods). To qualitatively assess the distribution of the features among participants, we visualized the structure of the latent space using t-SNE dimensionality reduction. Individuals with both higher BMI (Figure \ref{fig:figure4}B) and higher resting heart rate \ref{fig:figure4}C) were grouped closer together in the latent space, suggesting that the model effectively captured the combined influence of these factors on insulin resistance. To quantitatively validate this observation, we trained a classifier to predict “high BMI” and “high resting heart rate” directly from the learned representations. This classifier achieved performance of over 0.85 AUROC for both features, providing strong quantitative support for the qualitative insights derived from the t-SNE visualization. This ability to uncover complex interactions between key features, even in the absence of direct labels, highlights the potential of representation learning to reveal novel insights into the underlying physiology of insulin resistance.

\subsection{Stratified performance analysis of the insulin resistance prediction based on BMI and physical activity}
Because our cohort is heterogeneous in lifestyle, encompassing individuals with varying BMIs and levels of physical activity, we aim to assess the performance of subgroups based on their weight and activity levels. We categorized individuals into four groups based on BMI: obese (BMI $ \geq $ 30, 458/1165 participants, 39\% of the cohort), overweight (25 $ \leq $ BMI < 30, 378/1165 participants, 32\% of the cohort), normal weight (18.5 $ \leq $ BMI < 25, 319/1165 participants, 27\% of the cohort), and underweight (BMI < 18.5, 10 participants, 1\% of the cohort). Figure \ref{fig:figure4}D demonstrates that the majority of obese individuals in our study are insulin resistant (205/458 participants, 44\%). This supports the observation we showed in Figure \ref{fig:figure1}C, showing a significant positive correlation between BMI and HOMA-IR. We used the median daily step count to categorize individuals into five physical activity levels: sedentary (daily steps < 5000) (232/1165 participants, 20\% of the cohort), low active (5000 $ \leq $ daily steps < 7500, 376/1165 participants, 32\% of the cohort), somewhat active (7500 $ \leq $ daily steps < 10000, 313/1165 participants, 28\% of the cohort), active (10000 $ \leq $ daily steps < 12500, 133/1165 participants, 12\% of the cohort), highly active (daily steps $ \geq $ 12500, 100/1165 participants, 9\% of the cohort). Figure \ref{fig:figure4}E indicates that sedentary individuals are more likely to be insulin resistant (104/232 participants, 45\%). This aligns with the findings in Figure \ref{fig:figure1}C, which demonstrates a significant negative correlation between the number of daily steps and HOMA-IR. Supplementary Figure S5 shows the distribution of individuals with different obesity and fitness classes, where 147/1165 (12.6\%) of the cohort are obese and sedentary.

We then calculated the performance of predicting insulin resistance of each subgroup (Supplementary Figure S6), using sensitivity, specificity, and adjusted specificity. Adjusted specificity is a metric used in three-class classification systems to emphasize false positives at the extreme end of the healthy class \cite{sleep-apnea-filing}, insulin sensitivity in our case. In our context, false positives in adjusted specificity would be individuals predicted to be IR but who are actually IS (i.e., excluding those with Impaired-IS). The rationale behind this metric is that individuals with Impaired-IS who are predicted as IR would likely benefit from lifestyle interventions, as their bodies are progressing towards significant IR. Using a model incorporating wearables, demographics, and routinely available blood biomarkers, Figure \ref{fig:figure4}F demonstrates that IR predictions for obese participants (sensitivity = 0.86, adjusted specificity = 0.93) are superior to those for the entire cohort. Performance further improves when the model is tested exclusively on sedentary individuals (sensitivity = 0.88, adjusted specificity = 0.97). Our model exhibits superior performance in predicting IR for sedentary and obese individuals (sensitivity = 0.93, adjusted specificity = 0.95). This subgroup would likely benefit the most from lifestyle or therapeutic interventions aimed at improving insulin sensitivity, potentially reversing the course of diabetes or reducing the risk of future development.

\subsection{Robustness and consistency analysis of the insulin resistance prediction model }
To assess the consistency of our insulin resistance predictions over time, we performed a time-window sweep and computed predictions across different time windows (Methods) using both the "Wearables + Demographics" and "Wearables + Demographics+ Fasting Glucose" feature sets with the direct tree-based XGBoost model. For continuous HOMA-IR prediction, we calculated the coefficient of variation (CV) for each individual across all time windows as a measure of prediction stability. As depicted in Supplementary Figure S7A, the "Wearables + Demographics" model exhibited the highest variability, with a median CV of 9.85\% (interquartile range: 7.26\% - 13.04\%). Incorporating fasting glucose substantially improved prediction stability, reducing the median CV to 6.16\% (interquartile range: 4.32\%- 8.98\%) for the "Wearables + Demographics + Fasting Glucose" model. Similarly, incorporating the complete metabolic and lipid panels further reduced variations in predictions resulting in the median CV of 4.19 (interquartile range 2.89\% to 6.01\%), following the intuition that adding informative constant features stabilizes predictions across different time windows.

\begin{table}[]
\centering
\caption{Consistency of the proposed insulin resistance classifier measured for various seven-day windows for selected input feature sets.}
\label{tab:table_2}
\resizebox{\textwidth}{!}{%

\begin{tabular}{lccc}
\hline
\textit{\textbf{Classification Consistency}} & \textbf{Wearables + Demographics} & \textbf{\begin{tabular}[c]{@{}c@{}}Wearables + Demographics + \\ Glucose\end{tabular}} & \textbf{\begin{tabular}[c]{@{}c@{}}Wearables + Demographics + \\ Lipid Panel + Metabolic Panel\end{tabular}} \\ \hline
\multicolumn{4}{c}{\textbf{Contiguous 7-Day Windows}}                                                                                                                                                                                                                                    \\ \hline
\textit{100\% Consistency}                   & 74.32\%                           & 83.54\%                                                                                & \textbf{92.34\%}                                                                                             \\ \hline
\textit{75\% - 99.9\% Consistency}           & 7.39\%                            & 2.80\%                                                                                 & \textbf{1.91\%}                                                                                              \\ \hline
\textit{50\% - 74.9\% Consistency}           & 4.59\%                            & 3.31\%                                                                                 & \textbf{0.76\%}                                                                                              \\ \hline
\textit{\textless 49.9\% Consistency}        & 13.77\%                           & 10.33\%                                                                                & \textbf{4.97\%}                                                                                              \\ \hline
\multicolumn{4}{c}{\textbf{Contiguous 14-Day Windows}}                                                                                                                                                                                                                                   \\ \hline
\textit{100\% Consistency}                   & 81.76\%                           & 88.26\%                                                                                & \textbf{93.75\%}                                                                                             \\ \hline
\textit{75\% - 99.9\% Consistency}           & 4.59\%                            & 2.16\%                                                                                 & \textbf{1.27\%}                                                                                              \\ \hline
\textit{50\% - 74.9\% Consistency}           & 5.48\%                            & 4.33\%                                                                                 & \textbf{1.91\%}                                                                                              \\ \hline
\textit{\textless 49.9\% Consistency}        & 8.16\%                            & 5.22\%                                                                                 & \textbf{3.06\%}                                                                                              \\ \hline
\end{tabular}
}
\end{table}

To evaluate the consistency of insulin resistance classification, we examined the proportion of individuals whose predicted labels (IR vs. non-IR) remained stable across all time windows. Our results indicated that even the least stable model configuration achieved consistent and correct predictions for over 70\% of the individuals, regardless of the time window length or position within the study period (Table \ref{tab:table_2}). Focusing on the "Wearables + Demographics" model with the shortest time window (seven days), which intuitively should exhibit the highest variability, we observed consistent and accurate predictions for 74\% of the individuals. Furthermore, we investigated the characteristics of individuals whose predicted labels fluctuated across time windows. Intriguingly, these individuals were characterized by a change in resting heart rate and daily activity (represented by “daily steps”). This suggests that prediction instability might, in part, reflect genuine physiological fluctuations in insulin sensitivity rather than model limitations. To further explore the sources of variability, we analyzed the relationship between prediction stability (CV of HOMA-IR) and key factors such as time window length, individual feature values (e.g., average activity levels, BMI), and the presence of missing data. As expected, our analysis revealed that shorter time windows were associated with higher CVs, but more importantly, individuals with higher BMI values and resting heart rate exhibited more stable predictions. These findings provide valuable insights for optimizing model deployment and interpreting predictions in real-world settings, and collectively demonstrate the robustness of our models in predicting insulin resistance across 120 days, even when using only 7 days of data.

Supplementary Figure 7B illustrates a pattern among participants who do not exhibit 100\% consistency in their week-to-week insulin resistance prediction. For instance, participant \#13680 is IS, yet our model predicts them as IR for all weeks except two, where it erroneously predicts IS. Deeper analysis of the participant's weekly RHR, HRV, and step count revealed that on those two occasions predicted as IS, the participant had an outlier number of steps, likely placing them in an "active" region, thus leading to the model's error. This figure also highlights that by taking a majority vote over a longer prior period, such prediction errors can be minimized.

\begin{table}[]
\centering
\caption{Validation Cohort Characteristics.}
\label{tab:table_3}
\resizebox{\textwidth}{!}{%
\begin{tabular}{ccccccc}
\hline
\multicolumn{1}{l}{}                       & \multicolumn{1}{l}{}      & \textbf{ALL}    & \textbf{IS}      & \textbf{IMPAIRED-IS} & \textbf{IR}     & \textbf{P-Value}     \\ \hline
\textbf{\# of Participants}                & N                         & 72              & 33               & 20                   & 19              & \multicolumn{1}{l}{} \\ \hline
\multirow{2}{*}{\textbf{Demographics}}     & AGE                       & 44.5 (11.6)     & 43.7 (12.1)      & 44.4 (10.1)          & 45.9 (12.7)     & 0.8                  \\ \cline{2-7} 
                                           & BMI                       & 30.6 (9.0)      & 28.2 (7.3)       & 29.3 (6.3)           & 36.0 (11.9)     & 0.007                \\ \hline
\multirow{3}{*}{\textbf{Sex at Birth}}     & Female                    & 50              & 22               & 15                   & 13              & 0.81                 \\ \cline{2-7} 
                                           & Male                      & 22              & 11               & 5                    & 6               & \multicolumn{1}{l}{} \\ \cline{2-7} 
                                           & Choose not to answer      & 0               & 0                & 0                    & 0               & \multicolumn{1}{l}{} \\ \hline
\multirow{6}{*}{\textbf{Ethnicity}}        & Asian                     & 31              & 13               & 9                    & 9               & 0.21                 \\ \cline{2-7} 
                                           & Black or African American & 4               & 1                & 0                    & 3               & \multicolumn{1}{l}{} \\ \cline{2-7} 
                                           & White                     & 25              & 14               & 7                    & 4               & \multicolumn{1}{l}{} \\ \cline{2-7} 
                                           & Hispanic or Latino        & 5               & 1                & 2                    & 2               & \multicolumn{1}{l}{} \\ \cline{2-7} 
                                           & Mixed Race                & 6               & 4                & 2                    & 0               & \multicolumn{1}{l}{} \\ \cline{2-7} 
                                           & Choose not to answer      & 1               & 0                & 0                    & 1               & \multicolumn{1}{l}{} \\ \hline
\multirow{4}{*}{\textbf{Wearables}}        & RHR (bpm during sleep)    & 68.0 (7.5)      & 66.1 (6.5)       & 68.3 (6.7)           & 71.1 (9.0)      & 0.06                 \\ \cline{2-7} 
                                           & SLEEP DURATION (minutes)  & 423.5 (78.4)    & 423.7 (87.8)     & 434.7 (54.0)         & 411.5 (84.8)    & 0.66                 \\ \cline{2-7} 
                                           & STEPS (Daily)             & 8525.0 (4095.2) & 10070.5 (5184.2) & 7391.5 (1889.2)      & 7033.8 (2540.1) & 0.01                 \\ \cline{2-7} 
                                           & HRV (RMSSD during sleep)  & 39.3 (24.3)     & 42.3 (26.7)      & 40.9 (25.6)          & 32.3 (17.1)     & 0.35                 \\ \hline
\multirow{5}{*}{\textbf{Blood Biomarkers}} & HBA1C                     & 5.3 (0.8)       & 5.0 (0.5)        & 5.4 (0.9)            & 5.5 (1.1)       & 0.04                 \\ \cline{2-7} 
                                           & GLUCOSE                   & 92.7 (11.0)     & 87.8 (7.6)       & 91.7 (8.7)           & 102.1 (12.5)    & 0.0000083            \\ \cline{2-7} 
                                           & HDL                       & 51.7 (16.8)     & 57.1 (17.1)      & 50.8 (14.7)          & 43.2 (15.3)     & 0.01                 \\ \cline{2-7} 
                                           & LDL                       & 126.0 (35.4)    & 119.1 (31.9)     & 133.3 (35.6)         & 130.2 (40.5)    & 0.31                 \\ \cline{2-7} 
                                           & TRIGLYCERIDES             & 101.8 (50.0)    & 86.8 (41.1)      & 109.8 (53.3)         & 119.4 (55.4)    & 0.05                 \\ \hline
\end{tabular}
}
\end{table}

\subsection{Validation of the Proposed Insulin Prediction Models on an Independent Cohort}

\begin{figure*}
    \centering
    \includegraphics[width=0.95\textwidth]{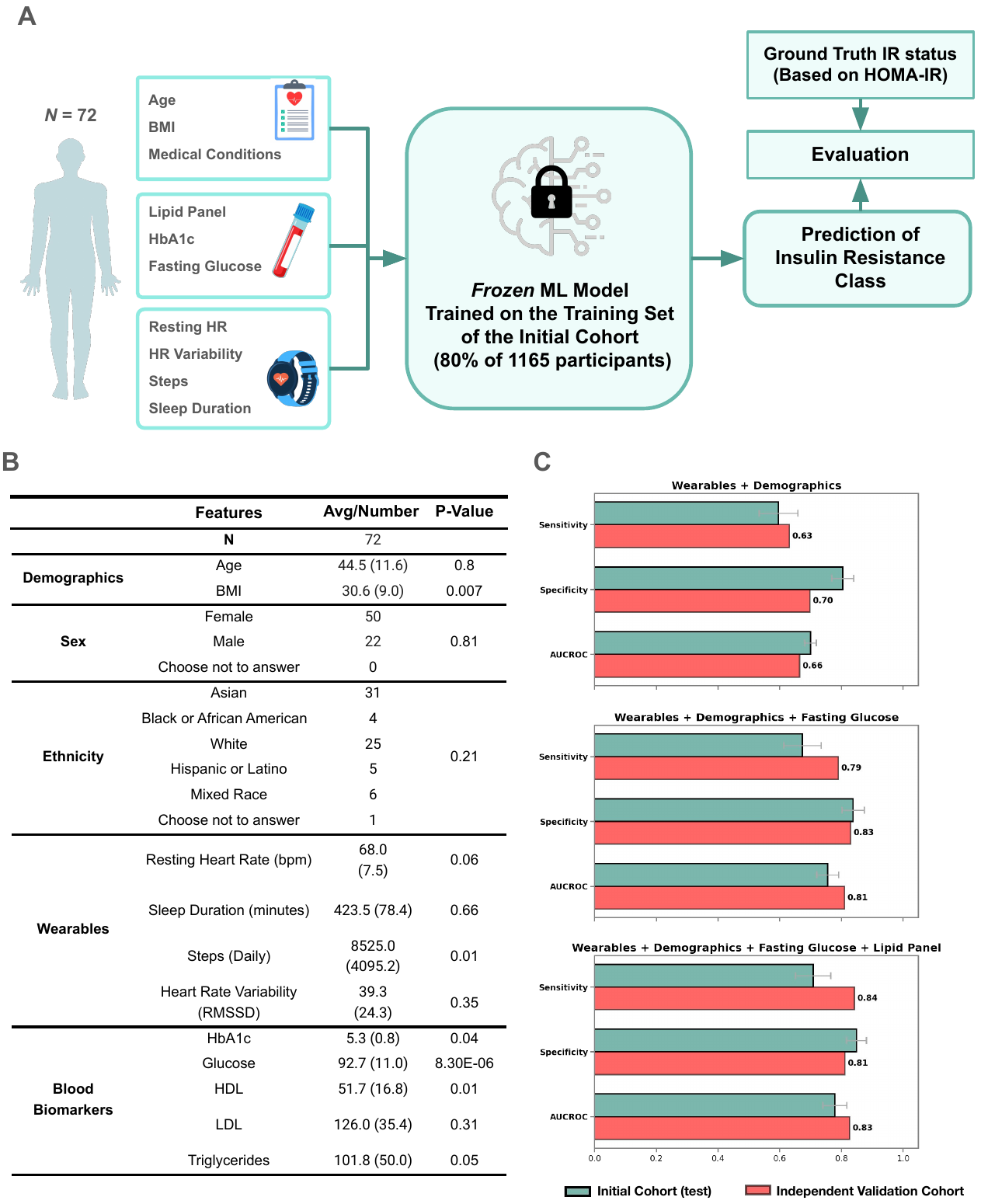}
    \caption{\textbf{Validation of Proposed Models on an Independent External Validation Cohort.} (A) Overview of the study for the independent validation cohort.  (B) Summary of the population characteristics of the external validation cohort. (C) Accuracy of selected ML models on the external validation cohort, including a side-by-side comparison with the results on the initial training cohort. We validated all of our trained models on this cohort except models with complete metabolic panel (CMP), since the external validation cohort did not include CMP in the blood tests.}
    \label{fig:figure5}
\end{figure*}

To validate the generalizability of the proposed insulin resistance models on unseen data, we evaluated the performance of our trained insulin resistance prediction model. The study protocol was approved by WCG (Institutional Review Board \#1371945) (Methods). Data collected throughout the study included anthropometric measurements (e.g., BMI, waist circumference, skin tone, and blood pressure), blood biomarkers (such as HbA1c, fasting glucose, fasting insulin, and lipid panel), wearable device data (Fitbit Charge 6, measuring heart rate, resting heart rate, heart rate variability, sleep duration, and step count), and health and lifestyle questionnaires (Figure \ref{fig:figure5}A). Initially, 144 individuals were enrolled, where 127 individuals had complete wearable data, and 82 individuals had complete physiological biomarker data acquired during an in-person visit at the end of the study. Ultimately, 72 individuals possessed both complete wearable data and complete physiological biomarker data, and this group was used as an independent validation cohort for the insulin resistance prediction model. This final cohort (N=72) was utilized to validate the generalizability of our insulin resistance prediction models. The cohort’ characteristics have an average age of 44.5 years, a BMI of 30.6 kg/m$^2$, and mixed ethnicities (Figure \ref{fig:figure5}B and Table \ref{tab:table_3}). Similar to the initial cohort (N=1165), HOMA-IR was used as the ground truth measure for insulin resistance in the independent validation study. Using HOMA-IR cut-off values of 2.9 and 1.5, the validation cohort was divided into three groups: 33 insulin sensitive, 20 with impaired insulin sensitivity, and 19 with insulin resistance. 

Using the frozen model learned from the initial cohort (80\% of the initial cohort), and using model with (Wearables + Demographic + Fasting Glucose + Lipid Panels), the insulin resistance prediction on the independent validation cohort performed on par with the test set of the initial study (Figure \ref{fig:figure5}C). Our frozen model achieved specificity of 0.81 (-0.04 from initial cohort, Experiment 0 in Table S9) and sensitivity of 0.84 (+0.15 with test set from the initial shown in Experiment 0 in Table S9). This shows the generalizability of the model on the new cohort. Supplementary Table S10 shows all evaluation metrics on the independent validation study for various modeling scenarios trained on the initial cohort.

\subsection{Insulin Resistance Literacy and Understanding Agent}

\begin{figure*}
    \centering
    \includegraphics[width=0.95\textwidth]{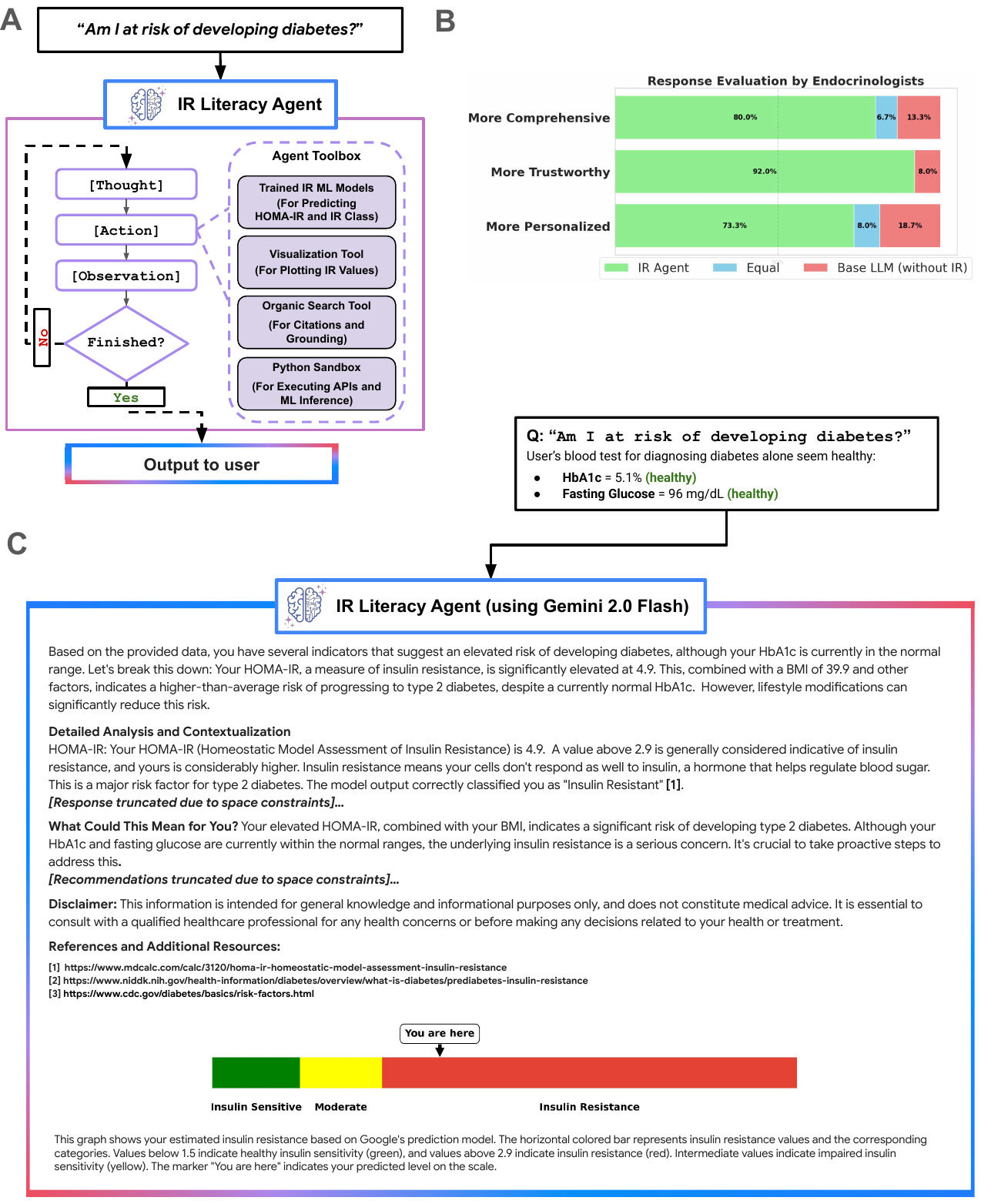}
    \caption{\textbf{Overview of Insulin Resistance Literacy and Understanding Agent (IR Agent). } (A) An illustration of the proposed IR agent (B) Win rate of our IR agent against the base model as evaluated by endocrinologists side-by-side (C) An example of a metabolically-relevant question paired with data from a real study participant, and the corresponding IR agent output.
}
    \label{fig:figure6}
\end{figure*}

The ability to detect insulin resistance from wearables and specific data points from personal health records before the onset of T2D (characterized by HbA1c > 6.5\%) raises the possibility of an Large Language Model (LLM) agent connected to users' wearables and personal health records notifying them of an increased risk of developing diabetes. It could also incorporate the inferred user's insulin resistance class interactively when a user asks general queries related to metabolic health. To show the potential of such a system for answering metabolically-relevant questions, we set out to design a reasoning agent (Methods), as illustrated in Figure \ref{fig:figure6}A. Our agent, called the Insulin Resistance Literacy and Understanding Agent (IR agent) utilizes a Reason and Act (ReAct) framework that is built on top of an LLM, in our case Gemini 2.0 Flash. Our agent combines the language understanding of an LLM with the ability to perform actions, such as searching the web for up-to-date information, accessing specialized tools like a calculator, and utilizing our novel HOMA-IR prediction models. This allows the IR agent to dynamically plan its response to a user's metabolic health query, grounding its answers in real-world data and verifiable calculations, rather than relying solely on the LLM's pre-existing knowledge. Upon receiving a user query, a dataframe of user's health data, along with the query itself, are provided to the agent within its context window, enabling the agent to tailor its reasoning and actions to the individual's specific health profile.

We evaluated efficacy and clinical relevance of the IR agent through a comprehensive assessment by five external, board-certified endocrinologists. This evaluation focused on two key aspects: (1) the added value of incorporating predicted IR information into an LLM response, and (2) the absolute accuracy and clinical safety of the IR Agent's outputs.  For the added value assessment, endocrinologists performed blinded side-by-side comparisons of responses generated by the IR Agent (with IR information) and a base LLM (Gemini 2.0 Flash, with access to the same user data but without IR prediction).  These comparisons, using queries from five representative study participants with diverse metabolic profiles (Supplementary Table S11), revealed a strong preference for the IR Agent across all evaluated dimensions as shown in Figure \ref{fig:figure6}B and Supplementary Table S12. Specifically, the IR Agent was rated as more comprehensive (80\% preference), trustworthy (92\% preference), and personalized (73.3\% preference) compared to the base LLM. Figure \ref{fig:figure6}C illustrates an example of query and response from the IR agent.

The absolute accuracy assessment involved endocrinologists evaluating the IR Agent's responses across four critical dimensions: factuality, data referencing and interpretation, safety, and grounding (citation validity). The IR Agent demonstrated high factuality (79\% of responses deemed completely factually accurate) and safety (96\% of responses considered safe) (Supplementary Figure S8A). A detailed inferred analysis of the data referencing and interpretation component revealed that our agent was able to accurately reference and interpret HOMA-IR values (100\% and 96\%, respectively) and demographic data (96\% for both referencing and interpretation). While referencing of wearables and blood biomarkers was consistently accurate, interpretation accuracy was lower for these data types (79\% and 59\% respectively; Supplementary S8B and Supplementary Table S13 for detailed breakdown), highlighting areas for refinement. Finally, 81\% of the citations provided by the IR agent were found to be relevant and verifiable. These expert evaluations provide evidence that incorporating predicted IR status enhances the quality, trustworthiness, and clinical utility of LLM-generated responses to metabolic health queries, forming a solid foundation for future development and deployment of similar AI-driven health assistants.

%% file: sections/3-discussion.tex
\section{Discussion}

Using a large and diverse cohort of 1,165 participants, we developed machine learning models to predict insulin resistance based on readily available lifestyle data (from wearables) and blood biomarkers that can be obtained from routine checkups. The models were trained using a validated ground truth measure of insulin resistance (HOMA-IR) that has been validated and established in previous large epidemiological studies \citep{Matthews1985-wg}. To classify individuals as insulin resistant (IR) or non-IR (based on a HOMA-IR threshold of 2.9), we first trained regression models to predict HOMA-IR values. Subsequently, we applied the 2.9 threshold to the predicted HOMA-IR values to assign individuals to the IR or non-IR categories. We preferred this regression-then-threshold approach over direct classification methods due to its enhanced interpretability, which allows for extraction of data patterns that facilitates further stratification and downstream analyses (Figure \ref{fig:figure2}B-E). We evaluated the HOMA-IR prediction models using regression coefficient of determination ($R^2$) and mean absolute error (MAE), while evaluating the predicted insulin resistance class using auROC, sensitivity, specificity, precision, and auPRC.

We tested 25 multimodal data combinations including wearables, demographics, fasting glucose, HbA1c, lipid panels, metabolic panels, and hypertension status. As expected, the model incorporating all modalities performed best. However, we identified three foundational models for predicting IR from wearables and personal health records, prioritizing ease of data collection alongside performance: (1) Minimal Model: Demographics and wearables lifestyle data. (2) Moderate Model: Wearables lifestyle data, demographics, and fasting glucose. (3) Optimal Model: Wearables lifestyle data, demographics, lipid panel, and metabolic panel (which include glucose). Demographic data is readily available, as is wearable data for users with smartwatches or trackers. Fasting glucose can be obtained from clinical labs or easily measured at home with a finger prick. Lipid panels and HbA1c are typically obtained from routine checkups, but are also accessible through microsampling devices and at-home testing kits (e.g., Mitra device (Neoteryx)).

Our deep learning framework for predicting insulin resistance represents the first deployable end-to-end model that utilizes readily available data from wearables, demographics, and routine blood biomarkers. In the United States, 26\% of the population owns smartwatches, and approximately 15\% undergo annual medical exams that include routine blood biomarker assessments. These figures are expected to increase with wider adoption of wearables and growing recognition of the health benefits associated with continuous health monitoring. 

In this work, we tested the hypothesis that insulin resistance can be estimated by measures of obesity (e.g., BMI, waist circumference, or body fat percentage) and fitness (e.g., resting heart rate, heart rate variability, step count, active zone minutes, or cardio load), and we showed that a model uses BMI, age, and wearables features (resting heart rate, heart rate variability, step count, and sleep duration) achieved 0.6 sensitivity, 0.81 specificity, and adjusted specificity of 0.94. The addition of blood biomarkers offers another orthogonal dimension, providing insights into actual physiology and metabolism. HbA1c and fasting glucose alone are not reliable markers for insulin resistance. 20\% (183/906) and 52\% of normoglycemic participants (HbA1c<5.7\%) exhibit IR and impaired-IS, respectively. This highlights the unique advantage of our models in the early prediction of insulin resistance in those at higher risk of developing type 2 diabetes, even if HbA1c and glucose are not yet elevated.

The gold standard test for insulin resistance, the euglycemic clamp \citep{Matthews1985-wg, Pei1994-oy, Greenfield1981-yn}, is impractical for routine clinical use due to its complexity and resource intensiveness.  Currently, HOMA-IR is not routinely assessed due to the cost and logistical challenges of insulin testing. While a single insulin measurement can be performed within reasonable cost in clinical settings, routine insulin assessment presents logistical (clinical lab visits) and cost-related challenges. At-home testing is unlikely to mitigate these logistical hurdles, as standard insulin immunoassays are not readily adaptable to user-friendly, home-based microfluidic kits yet. Consequently, repeated insulin testing necessitates ongoing clinical lab visits, leading to significant cumulative costs. Therefore, our model that uses readily available digital and blood biomarkers could serve as a screening tool to prioritize individuals for testing insulin at clinical lab settings to calculate the exact HOMA-IR.

Previous research involving wearable smartwatches and insulin resistance primarily focused on investigating associations between wearable-derived features (e.g., resting heart rate, heart rate variability) and HOMA-IR, without developing predictive models \citep{Saito2015-ib, Svensson2016-ts, Poon2020-cj, Flanagan1999-sa, Beddhu2009-yw, Saito2022-fk, Grandinetti2015-ql}. The most recent state-of-the-art method for detecting insulin resistance at-home involves performing two oral glucose tolerance tests (OGTTs) at home using continuous glucose monitoring (CGM) to analyze glucose time series data. These methods have demonstrated an impressive AUROC of 0.88 \citep{Metwally2024-cgm}. While the ability to infer IR is a crucial step towards developing widely accessible diagnostic tests, the adoption of CGMs among the non-diabetic population remains limited. Therefore, our approach, which leverages wearable smartwatch devices, demographic data, and readily available blood markers, offers a more scalable solution that does not require additional testing. Regarding blood biomarkers, while numerous studies have demonstrated strong correlations between HOMA-IR and biomarkers such as triglycerides and HDL \citep{Simental-Mendia2014-th, diabetes-clinical-practice, Hirschler2015-at, Olson2012-ul}, few have aimed to predict insulin resistance using readily available blood markers. Existing models either exhibited suboptimal predictive performance, relied on small sample sizes, or included insulin itself as a predictor—a marker not typically included in routine annual exams \cite{McAuley2001-ri}.

While many studies have developed predictive models for insulin resistance using demographic and clinical biomarkers \citep{Araujo2023-sd, Berglund1996-jm, Park2022-kq, Wedin2012-pt}, the integration of wearable data remained unexplored. Individual lifestyle factors, such as physical activity, significantly influence insulin resistance, making the potential of wearable devices to enhance prediction a crucial area of study. Our study demonstrates that, although wearable data may not significantly improve overall population-level prediction compared to models using demographics alone, it objectively identifies sedentary individuals based on average daily step count. Figure \ref{fig:figure4}F illustrates this stratification for our optimal model, showing a sensitivity of 88\% on the sedentary group compared to 76\% in the overall population, while maintaining adjusted specificity of 97\%. This targeted approach highlights the utility of wearable data for enhancing diagnostic precision in high-risk individuals.

Moreover, prior studies utilized anthropometric measures (e.g. BMI, waist circumference, etc) or lifestyle surveys (e.g., activity and dietary and physical activity) to predict insulin resistance. However, these methods generally exhibited suboptimal performance, primarily because they failed to capture underlying physiology or provide continuous monitoring of lifestyle factors \citep{Wedin2012-pt, youth-pediatrics}. Notably, \cite{Araujo2023-sd} proposed logistic regression models for IR prediction, utilizing anthropometric measures and blood biomarkers45. Employing a pediatric cohort from Portugal and defining IR using a HOMA-IR threshold of 3.4 (the 95th percentile in healthy Mediterranean children and adolescents), their model, which incorporated BMI, obesity duration, brachial, waist, and hip circumferences, acanthosis nigricans, Tanner stage, self-reported physical activity, family history of type 2 diabetes, hypertension, and fasting glucose, achieved a sensitivity and specificity of 0.816 in the training cohort. However, the validation cohort exhibited a sensitivity of 0.81 and a specificity of 0.52. In contrast, our model, utilizing demographics (only BMI and age), passively and objectively measured physical activity via wearables features, and fasting glucose, achieved a specificity of 0.84 and a sensitivity of 0.73 in the initial cohort, and a specificity of 0.83 and a sensitivity of 0.79 in the independent validation cohort. Waist circumference emerges as the top anthropic metric that is associated with insulin resistance. Waist index (WI), calculated as waist circumference (cm) divided by 94 for men and 80 for women, has demonstrated a strong correlation with HOMA-IR \citep{MAGRI2016}, and it was used to predict IR (HOMA-IR>2.5), with sensitivity of 0.78 and specificity of 0.65. In our study, we initially used a HOMA-IR threshold of 2.9, which yielded sensitivity of 0.78 and specificity of 0.84. However, upon recalculating sensitivity and specificity using a threshold of 2.5, we obtained a sensitivity of 0.82 and specificity of 0.78. Our method provides a scalable, initial screening tool to identify individuals with an increased likelihood of insulin resistance. Positive screening results would prompt referral for clinical fasting glucose and insulin assays, allowing accurate HOMA-IR calculation and subsequent clinical discussion. Consequently, a model with higher specificity than existing surrogates is essential to minimize unnecessary testing in laboratory settings. 

Multiple studies have associated lifestyle intervention programs, such as weight loss and increased aerobic exercise, with significant reductions in insulin resistance \citep{McLaughlin2008-xp, Ryan2020-cm}. Future longitudinal studies should validate whether weight loss in individuals leads to decreases in insulin resistance detectable by wearable devices. In addition to longitudinal validation, simulations could model the magnitude of lifestyle changes needed to improve insulin resistance and the degree of change required to reverse the progression of diabetes\cite{Park2024-lu}.

While our study included diverse participants in terms of geography, age, gender, and other factors, only 25\% of participants had complete data and were included in the analysis; the final cohort may have overrepresented those with cardiometabolic disease or health awareness due to the requirement for blood lab tests and use of wearable devices. Moreover, genetics and the microbiome are established contributors to insulin resistance \citep{Zhou2019-vx, Sailani2020-rb, Hornburg2023-ie, Udler2019-ye, Mahajan2022-fk, Contrepois2020-at}. Future studies would benefit from collecting and analyzing such data in relation to insulin resistance prediction models. This will elucidate the interplay between lifestyle and genetic factors in insulin resistance. Regarding ethnicity, although the Caucasian demographic was predominant, we conducted stratified analyses to assess the model's performance across different ethnic groups. Specifically, we examined whether a model trained primarily on Caucasian data would generalize to Hispanic, East Asian, Indian, African American, and Native American participants (Supplementary Table S9). Our findings indicated that the model performed similarly for Caucasian, Hispanic, and East Asian groups. This suggests that the current training set allows for reasonable generalization to these populations. This observation was also supported by the independent validation cohort. However, we observed that the model did not generalize effectively to Indian, African American, or Native American participants. One potential explanation for this lack of generalization is the limited representation of these ethnicities within our training cohort. It is possible that the physiological characteristics of these groups require a larger and more diverse training set to be adequately captured. Therefore, we propose that future research should focus on developing ethnicity-specific models for African American and Indian populations to improve insulin resistance prediction. Similarly, to enhance the model's sensitivity for predicting insulin resistance in less vulnerable populations, such as individuals with normal BMI and active lifestyles, we recognize the need for larger cohort sizes. These mixture of expert models and expanded datasets will be crucial for ensuring equitable and accurate predictions across diverse populations.

Moreover, we recognize that the current adoption of advanced wearables is skewed towards Caucasian. However, many basic wearable fitness trackers with the necessary measurement capabilities are becoming increasingly affordable and widely available, and we expect this trend to continue. In our models, we incorporated features commonly found on lower-cost devices, such as resting heart rate, daily steps, sleep duration, and heart rate variability. Additionally, recognizing the limitations of affordable wearables in providing high-resolution data, we utilized aggregate measures of these metrics rather than time-series data, ensuring the applicability of our models to a wider range of devices while maintaining their relevance and practical utility. We acknowledge that consumer smartwatch brands, sensors, and firmware may provide different accuracy, precision, and temporal resolution for measuring digital biomarkers (heart rate, steps, sleep) compared to research-grade or clinical-grade devices. In our study, we utilized data from seven different smartwatches and four trackers, all from the same manufacturer, Google/Fitbit, and employing similar algorithms (Supplementary Figure S2). We implemented extensive quality control measures to ensure the data used for the model were of high fidelity and accuracy, and that the model received sufficient wearable data input before generating a prediction (Supplementary Figure S1). While we expect the results to generalize to other brands, further investigation and more specific quality control steps to mitigate known accuracy issues are warranted. Furthermore, Table \ref{tab:table_2} demonstrates that aggregating these features within our model over longer periods (7 days vs 14 days) reduces the potential noise introduced by acute lifestyle changes (Supplementary Figure S7).

Our training of the machine learning models relied on HOMA-IR as a proxy to scale the experiments to thousands of participants instead of relying on the gold standard euglycemic clamp which is laborious and costly. However, using HOMA-IR as a ground truth presents some challenges, primarily related to per-subject reproducibility and standardization of insulin concentrations across laboratories. This can lead to a reported coefficient of variation of 23.5\% between two measurements \citep{Sarafidis2007-zw}. In our study, all participants underwent blood tests at Quest Diagnostics branches across the United States. This approach minimizes the risk of variations in insulin measurement standardization across different laboratories, and hence less HOMA-IR variability. While acknowledging this limitation, we emphasize that HOMA-IR remains a valuable tool for classifying individuals into broad categories of insulin sensitivity or resistance, rather than providing precise quantifications of resistance levels. Therefore, our framework prioritizes the evaluation of insulin resistance class over the prediction of specific HOMA-IR values. Improved standardization of insulin measurements will further mitigate this issue.
In the future, insulin resistance detection could be employed as a more objective measure for prescribing GLP-1 and GIP agonists. Currently, providers rely on BMI guidelines to determine the need for these medications \citep{European_Medicines_Agency_undated-aq}, which is not an ideal metric to differentiate between those who require the drug for medical necessity versus those who may be using it for lifestyle purposes. Utilizing an IR measure would provide both providers and payors with a more robust and objective assessment when prescribing and reimbursing for these drugs. However, the proposed insulin resistance detection models would need further fine-tuning to improve sensitivity and specificity, particularly on the sub-population that would benefit most from incretin hormone-based therapies.

%% file: sections/4-methods.tex
\section{Methods}
\label{sec:methods}

\subsection{Remote participant recruitment via
GHS}\label{remote-participant-recruitment-via-ghs}

The study protocol was approved by the institutional review board at
Advarra (Columbia, MD, Protocol Pro00074093). All participants provided
informed consent. In addition to the consent, participants were also
asked to sign the Quest HIPAA authorization as part of the consent flow
within the GHS app. Participants were recruited using the ``Google
Health Studies'' (GHS) Application and Fitbit App. GHS is a platform to
run digital studies that allows participants to enroll into studies,
check eligibility and provide informed consent. GHS enables the
collection of wearable features using Fitbit devices or Pixel watches
and allows participants to complete questionnaires and order blood tests
with Quest Diagnostics. This prospective, observational study resulted
in enrollment of 4416 participants in the United States, where a subset
of those (N=1165) had complete data and thus were included in our
analysis (\textbf{Supplementary Figure S1}). The study was conducted
remotely with one visit to a Quest Patient Service Center for a blood
draw. Participants were asked to continuously wear their wrist-worn
wearable devices, schedule and complete a blood draw, and answer
questionnaires. This study used an electronic informed consent process
managed by the GHS App using electronic signatures.

Inclusion criteria included U.S. residents between the ages of 21 - 80,
Fitbit users in Android that wear a Fitbit wearable device or a Pixel
Watch with heart rate sensing capabilities, users who have at least 3
months of existing data where they have used the device for at least
75\% of the days to track their activity and sleep, participants willing
to update their Fitbit Android App, participants willing to install or
update their Google Health Studies (GHS) app on their Android phone,
participants willing to link (or create) their Quest Diagnostics account
to the GHS app, participants able to speak and read English and provide
informed consent and HIPAA authorization to participate in study,
participants who have access to and be willing to go to a Quest
Diagnostics location for blood draws. Wearable devices were limited to
any model of a Fitbit device or Pixel watch with heart rate (HR) sensing
capabilities (e.g., Pixel Watch, Charge HR, Charge models 2 - 5, Sense,
Sense 2, Versa, Versa 2, 3 \& 4, Versa Lite, Inspire HR, Inspire 2, and
Luxe).

Exclusion criteria included participants living in Alaska, Arizona,
Hawaii and the US territories since Quest Diagnostics cannot provide
blood tests in those states, participants with uncontrolled disease (for
example, a recent change in treatment in the past 6 weeks, awaiting
review to trigger a change of treatment, a treating physician has
indicated the condition is not yet controlled, or where symptoms of the
condition are not responding to treatment), or participants with
conditions that might make collection of blood samples through
venipuncture impractical.

\subsection{Study design}\label{study-design}

As part of this study, participants were asked to link their Fitbit
account with the GHS app. They were also asked to grant GHS permission
to collect Fitbit data throughout the study, including data for up to 3
months before study enrollment. Once participants were enrolled in the
study they were asked to perform the following: (1) wear their Fitbit
device or Pixel watch during the day and while they sleep (at least 3
out of every 4 days) for the duration of the study; (2) complete four
questionnaires which will request demographic information, health
history and health information such as sleep and exercise habits,
participant perception of health, and blood test interpretation (see
below for details); (3) schedule an appointment to complete the lab test
orders that were placed for them and go to a Quest Patient Service
Center for a blood draw within 65 days of enrollment; (4) complete a
blood draw at the Quest Patient Service Center; (5) review blood test
results in the GHS App when available.

\subsection{Metadata collection}\label{metadata-collection}

Demographics (e.g., age, gender, ethnicity, weight, height), optional
measures, such as medical conditions (e.g. diabetes, hyperlipidemia,
cardiovascular disease, kidney disease, hypertension, etc.), blood
pressure, waist circumference, medications, self-reported health
management and habits were collected through a baseline questionnaire
that participants completed immediately after enrollment through the GHS
app.

\subsection{Blood biomarkers via Quest
diagnostics}\label{blood-biomarkers-via-quest-diagnostics}

Eligible participants were asked to schedule and complete a visit at a
Quest Diagnostics Patient Service Center of their choice in their local
area. This visit included a standard blood draw to measure the
following: Complete Blood Count, Comprehensive Metabolic Panel, insulin,
total cholesterol, triglycerides (TG), HDL-cholesterol, calculated
LDL-cholesterol ``calculated'', HbA1c, hs-CRP, GGT and Total
Testosterone. For this research study, we only had access to this
predefined set of lab tests for this study and did not collect or
receive any blood test data not included in this study. Participants
were asked to have their blood drawn while fasting for at least 8 hours,
early morning, 7am-10am local time, to minimize the effect of the solar
diurnal cycle. This study also had clinical oversight by a physician
network partner. Lab results from the blood draw were returned to
participants and made available for participant review in the GHS app
for the duration of the study; however, these were pulled directly from
Quest Diagnostics each time it is requested by the participant and not
stored in the GHS app. Participants provided consent and HIPAA
authorization to grant GHS permission to collect the corresponding
results from Quest Diagnostics. Data transferred from Quest Diagnostics
were retrieved securely using encrypted protocols. Once the study was
completed, lab results remained in the participant's Quest Account in
accordance with Quest\textquotesingle s standard practices.

\subsection{\texorpdfstring{Wearables data via Fitbit devices and Pixel
Watches
}{Wearables data via Fitbit devices and Pixel Watches }}\label{wearables-data-via-fitbit-devices-and-pixel-watches}

Participants were asked to wear their own Fitbit device or Pixel Watch
that may include sensors such as PPG optical heart rate tracker,
multipurpose electrical sensors, gyroscope, altimeter, accelerometer,
on-wrist skin temperature, blood oxygen saturation (SpO2), and ambient
light sensor. Participants consented to share the following data from
these devices: (1) heart rate metrics: heart rate (HR), resting heart
rate captured daily, interbeat interval (IBI, also known as RR interval)
calculated from the PPG sensor, and HRV metrics (such as RMSSD, SDNN,
SDRR, pNN50, etc.) (2) physical activity metrics: Measures of physical
activity, including steps, floors, and active zone minutes (AZMs), (3)
sleep metrics: bedtime, wake time, sleep duration, sleep stages, sleep
quality, sleep coefficients, and sleep logs, (4) respiration and skin
temperature: Respiration rate values during the day/night and skin
temperature (if the sensor is available on the wearable), (5) SpO2:
Blood oxygen saturation values during the day/night (if available on the
wearable), (6) weight: Measure of weight that may be logged in the
Fitbit account, (7) exercise and activity data: Daily total exercise
sessions completed and logs of activities that have been classified.
Fitbit daily RHR is calculated from periods of stillness throughout the
day, as determined by the on-device accelerometer. If a person wears
their device while sleeping, their sleeping heart rate is also included
in the calculation. Additional details are described in \cite{Russell2019-qg, Speed2023-qp}. Daily RMSSD HRV is calculated from pulse intervals measured during sleep
periods greater than 3 hours \citep{Fitbit_heart_rate_tracking}. AZMs
is a feature that tracks the time a person spends in different heart
rate zones during physical activity. A person receives 1 AZM for every
minute spent in the ``moderate'' zone, and 2 AZMs for every minute spent
in the ``vigorous'' or ``peak'' zones \citep{Fitbit-azms}. The
heart rate zones are based on percentage of heart rate reserve achieved,
where the heart rate reserve is the difference between maximum heart
rate and resting heart rate Moderate zone is defined from 40-59\% of
heart rate reserve, vigorous is from 60-84\%, and peak is 85-100\% \citep{Fitbit_heart_rate_tracking, Fitbit-azms2}.

\subsection{Selection of HOMA-IR thresholds for Insulin
Resistance}\label{selection-of-homa-ir-thresholds-for-insulin-resistance}

Our selection of HOMA-IR thresholds was based on \citep{Gayoso-Diz2013-pf}, which
reported a HOMA-IR range of 1.5 to 3 for insulin resistance. We chose a
threshold of 2.9, approximating the midpoint between the NHANES-derived
threshold for the U.S. population (2.77) and the maximum value in the
review (3). For insulin sensitivity, we used the lowest threshold from
the review, 1.55, rounded down to 1.5. Participants with HOMA-IR values
between 1.5 and 2.9 were classified as having impaired insulin
sensitivity (impaired-IS).

\subsection{Modeling and Computational
Pipeline}\label{modeling-and-computational-pipeline}

As shown in \textbf{Figure 1B}, our method consists of four stages: (i)
Data preprocessing, (ii) modeling and training, (iii) Prediction and
classification of HOMA-IR, and (iv) LLM-based interpretation of the
results. We describe each of these components, including evaluation
strategies and large-scale ablation studies, below.

\subsubsection{Data preprocessing}\label{data-preprocessing}

\textbf{Age and Body Mass Index (Demographics):} From the user-provided
data on recruiting surveys, we extract a user's age and compute BMI from
height and weight. As a quality control process, we exclude users with a
BMI greater than 65, or smaller than 12.

\textbf{Digital Markers Derived from Wearables (Wearables Features):}
Using the estimated digital markers from Fitbit algorithms, we
aggregated users' digital markers using mean, standard deviation and
median values for the past \(n = \{ 7,\ 14,\ 30,\ 60,\ 90,\ 120\}\) days
prior to blood test collection. We performed ablation studies to find
the ``optimal'' value of \(n\) (described later in this section).

\textbf{Biomarkers from Blood Biochemistry (Blood Tests):} As a first
filter, we removed any participant who was not fasting as reported in a
survey at the time of blood test collection. Additionally, for each
experiment, we exclude participants with missing values from the input
feature set. Lastly, to remove outliers, we use the true HOMA-IR value
(HOMA-IR = {[}Fasting Insulin (µU/ml) × Fasting Glucose (mg/dL) {]} /
405) and exclude any participant whose HOMA-IR value is greater than or
equal to 15.

\textbf{Data Standardization:} The data used for modeling is a
concatenation of demographics, aggregated digital markers and blood
biomarkers. In order to create a consistent modeling data that is
agnostic to the learning model, we standardized input features to have
zero mean and unit variance. For each training fold, our ``normalizer''
object was fit to the data in the training subset (not including samples
in the testing subset). The fitted object was then used to transform the
samples in both the training and testing subsets. The standardized data
was used for all modeling tasks and evaluations.

To test the generalizability of our approach to the validation cohort,
we used the same ``normalizer'' object that was fitted to the training
data from the initial cohort (the model used in Experiment 0 in Table
S9).

\subsubsection{\texorpdfstring{Modeling }{Modeling }}\label{modeling}

In order to determine risk of insulin resistance, our goal is to first
predict the value of HOMA-IR, and use existing thresholds for
classifying individuals. Performing regression prior to classification
allows for flexibility and much greater interpretation of our results
and analysis of individual data points.

\textbf{Direct Regression}

As our first approach in modeling HOMA-IR, we utilize Gradient Boosting
Machines, specifically the XGBoost framework \citep{Chen2016-hp}.
Gradient boosting methods excel in handling complex datasets with
potentially non-linear relationships, making them particularly
well-suited for our task. Compared to traditional tree-based methods,
XGBoost provides computational efficiency, scalability, and,
importantly, regularization techniques that enhance model
generalization. While often associated with tree-based models, XGBoost
offers a versatile framework for leveraging weak learners, which include
both linear and tree-based learners. Due to the ambiguity surrounding
the linear interaction between various input features, we utilize both
linear (\emph{L1-L2}) and nonlinear (\emph{tree-based}) learners to
assess the complexity of the problem space.

Given \(n\) data-label pairs \((x_{i},\ y_{i})\), gradient boosting can
be viewed as an additive combination of \(K\) weak learners \(f\), with
the aim to predict target \({\widehat{y}}_{i}\) through:

\begin{equation}
    \widehat{y}_{i} = \sum_{k}^{K}f_{k}(x_{i}),\ f_{k} \in \mathcal{F}, 
\end{equation}

\(\mathcal{F}\) denotes the space of all possible regression trees,
\(\mathcal{F} = \{ f(x)\  = \ w_{q}(x)\}\) with
\(w \in {\mathbb{R}}^{T}\)where \(T\) is the number of leaves in the
tree, and \(q:\ {\mathbb{R}}^{d} \rightarrow T\) represents the
structure of each tree that map \(d\)-dimensional data points (i.e.
\(x_{i} \in {\mathbb{R}}^{d}\)) to the corresponding leaf index. In our
notation, \(f_{k}(x_{i})\) represents the prediction made by the
\(k\)-th tree in the ensemble for the input sample \(x_{i}\). While the
choice of learners \(f\) can determine the linearity of the model, the
number of trees and number of leafs serve as important hyperparameters
that can further control model complexity. The objective of extreme
gradient boosting machines is to minimize the loss function

\begin{equation}
\mathcal{L}(\phi)  = \sum_{i}^{n}l({\widehat{y}}_{i}, y_{i})  + \sum_{k}^{K} \Omega(f_{k}),
\end{equation}

Where \(l({\widehat{y}}_{i},\ y_{i})\) represents the (traditional
gradient boosting machines) loss function measuring the discrepancy
between the predicted output \(\widehat{y_{i}}\ \)and \(y_{i}\), and the
regularizer \(\Omega(f_{k})\) penalizes model complexity, where
\(\Omega = \gamma T + \ \frac{1}{2}\lambda\| w\|^{2}\) with \(\gamma\) and
\(\lambda\) being hyperparameters.

In our work, the first set of models use weak linear learners which can
capture additive linear relationships within the data. This approach
assumes that the target variable can be modeled as a linear combination
of the input features, potentially revealing important feature
interactions (this is obtained by using \texttt{gblinear} from the XGBoost
implementation \citep{xgboost-implementation}. Recognizing the potential for complex nonlinear relationships within our
data, we let the second set of models to be nonlinear (through \texttt{gbtree} in
the XGBoost implementation). XGBoost\textquotesingle s efficient tree
construction algorithm, based on a pre-sorted split finding algorithm
and sparsity-aware split finding, allows for faster training and
exploration of a larger parameter space. This, coupled with the inherent
ability of decision trees to model nonlinear interactions, make the
tree-based approach a powerful technique for uncovering potential
complex patterns in our study.

\textbf{Representation Learning}

While direct regression methods, such as XGBoost, can effectively handle
high-dimensional data, we hypothesized that their performance can be
further amplified by providing informative and concise data
representations. Recent studies have demonstrated the importance of
mathematical representation of personal health records (blood tests,
lifestyle data, etc.) for various downstream tasks \citep{Heydari2024-tv, Heumos2023-dc, Zhou2023-el}. To test the representation learning hypothesis, we used representation learning techniques, specifically autoencoders (AEs) and masked autoencoders (MAEs), potentially learning latent representations that could enhance regression performance. We provide an overview of each
approach as well as implementation details below.

\emph{\textbf{Simple Autoencoders.}} As a first approach, we employ a
simple autoencoder (AE) architecture consisting of fully-connected
layers for both the encoder and decoder. This architecture, while
simpler than more complex variants, provides a robust baseline for
assessing the value of learned representations. In general, autoencoders
are a non-trivial mapping that can reconstruct the input, i.e.
\(AE(x_{i}) = \ \widetilde{x_{i}} = \ Dec(Enc(x_{i}))\), where
\(Enc( \cdot ):{\mathbb{R}}^{d}\  \rightarrow {\mathbb{R}}^{z}\ \)
denotes a mapping of input data onto a latent space, and
\(Dec( \cdot ):{\mathbb{R}}^{z}\  \rightarrow {\mathbb{R}}^{d}\)
represents the mapping from the latent space back to the input space. To
ensure accurate reconstruction of the inputs, traditional AEs aim to
minimize a mean squared loss function:

\begin{equation}
    \mathcal{L}_{AE}(x)  = {\mathbb{E}}_{x_{}}\| Dec(Enc(x)) - x\|^{2}.
\end{equation}

This encourages the model to extract the most informative features from
the input or to produce richer representations necessary for accurate
reconstruction. To further improve model learning, we also added a
\emph{smooth L1 loss} to the traditional AE loss in order to make it
more robust to outliers \citep{Girshick2015-lm} This
resulted in the objective function shown in Eq. (4).
\begin{equation}
    \mathcal{L}_{AE}(x)  = {\mathbb{E}}_{x_{}}\| Dec(Enc(x)) - x\|^{2} + \lambda \cdot SL(x, Dec(Enc(x)))
\end{equation}

where the smooth L1 Loss (\(SL\)) is defined as
\begin{equation}
    SL(x_{1}, x_{2})  = 
    \begin{cases}
    0.5 \lbrack x_{1}  - x_{2}\rbrack^{2}, \hspace{0.5cm} &\text{if   } |x_{1} - x_{2}|  < 1;\\
    |x_{1}  - x_{2}|  - 0.5,\hspace{0.5cm}&\text{otherwise}
    \end{cases}
\end{equation}

\emph{\textbf{Masked Autoencoder.}} While traditional AEs have been
shown to be an effective unsupervised approach for representation
learning, Masked Autoencoders (MAEs) \citep{He2022-ds}, a
self-supervised variant of AEs, have shown tremendous improvements over
traditional AEs, achieving state-of-the-art results across different
benchmark datasets \citep{He2022-ds, Zhang2022-rh}.
For our approach, given a datapoint \(x_{i}\  \in {\mathbb{R}}^{d}\), we
start by drawing a masking vector \(m\  \in {\mathbb{R}}^{d}\) from a
multivariate Bernoulli distribution with probability \(p\) (determined
through ablation studies, as discussed later). Using the mask vector, we
generate a masked version of the input,
\(\widehat{x_{i}} = x \odot m\ \), which is then inputted to the MAE
model. Similar to the traditional AE, the objective of our network is to
reconstruct the initial input, but from the masked vector
\({\widehat{x}}_{i}\) as opposed \(x_{i}\). Therefore, our object for
the MAE can be expressed as:
\begin{equation}
    \mathcal{L}_{AE}(x)  = {\mathbb{E}}_{x_{}}\| Dec(Enc(x \odot m)) - x\|^{2} + \ \lambda \cdot SL(x,\ Dec(Enc(x \odot m)))
\end{equation}

\textbf{\emph{Model Training and HOMA-IR Prediction from Learned
Representations.}} Results presented for regression in the main
manuscript include two stages: (i) representation learning, and (ii)
using the learned representation to train a XGboost with linear learners
for predicting HOMA-IR values. We chose using linear XGBoost models for
the second stage to better showcase the improvements made possible by
our representation learning approach.

We trained both models for a total of 500 epochs with the Smooth L1 loss
coefficient \(\lambda = 0.01\). For optimization, we used the Adam
optimizer with \((\beta_{1},\ \beta_{2})\  = \ (0.9,\ 0.999)\) and
\(\epsilon = 10^{- 12}\). The initial learning rate was set to
\(lr = 0.001\), scheduled to decrease exponentially (\(\gamma = 0.95\))
after 100 epochs on a plateau with a patience of 20 epochs. For the MAE
model, we set the masking probability \(p = 0.75\) based on our
empirical results (found through grid search) and existing literature.

\textbf{Parameter Selection.} To identify the set of parameters used for
the ablation studies, we performed grid search for key parameters. For
all models, due to grid search's exponential computational complexity,
we performed grid search on a common feature set and time window with
5-fold cross validation to identify optimal hyperparameter values. We
describe the parameter grid for each model below. To make our
experiments deterministic, our random seed was set to {[}0, 92, 1, 2024,
12121{]}.

\begin{table}[htbp] 
\centering 
\caption{Hyperparameter grids for XGBoost models.} 
\resizebox{\textwidth}{!}{
\begin{tabular}{lll}
\toprule
\multicolumn{1}{c}{Model}                     & \multicolumn{1}{c}{Hyperparameter} & \multicolumn{1}{c}{Grid}                                                                      \\ \hline
\multirow{6}{*}{XGBoost with Linear Learners} & \texttt{n\_estimators}                      & {[}5, 10, 15, 25, 50, 85, 100, 125, 150, 200{]}                                               \\ \cline{2-3} 
                                              & \texttt{learning\_rate}                     & {[}0.01, 0.05, 0.09, 0.1, 0.15, 0.19, 0.21, 0.25, 0.29, 0.31, 0.35, 0.39, 0.41, 0.45, 0.51{]} \\ \cline{2-3} 
                                              & \texttt{reg\_lambda}                        & {[}0.0, 0.1, 0.2, 0.3, 0.4, 0.5, 0.6, 0.7, 0.8, 0.9, 1{]}                                     \\ \cline{2-3} 
                                              & \texttt{reg\_alpha}                         & {[}0.0, 0.1, 0.2, 0.3, 0.4, 0.5, 0.6, 0.7, 0.8, 0.9, 1{]}                                     \\ \cline{2-3} 
                                              & \texttt{booster}                            & {[}"gblinear"{]}                                                                              \\ \cline{2-3} 
                                              & \texttt{random\_state}                      & Set to random seed                                                                            \\ \midrule
\multirow{7}{*}{XGBoost with Tree Learners}   & \texttt{n\_estimators}                      & {[}50, 100, 200{]}                                                                            \\ \cline{2-3} 
                                              & \texttt{learning\_rate}                     & {[}0.01, 0.1,0.15, 0.2, 0.3, 0.4, 0.5{]}                                                      \\ \cline{2-3} 
                                              & \texttt{reg\_lambda}                        & {[}0.0, 0.1, 0.2, 0.3, 0.4, 0.5, 0.6, 0.7, 0.8, 0.9, 1{]}                                     \\ \cline{2-3} 
                                              & \texttt{reg\_alpha}                         & {[}0.0, 0.1, 0.2, 0.3, 0.4, 0.5, 0.6, 0.7, 0.8, 0.9, 1{]}                                     \\ \cline{2-3} 
                                              & \texttt{booster}                            & {[}"gbtree"{]}  
                                              \\ \cline{2-3} 
                                              & \texttt{random\_state}                      & Set to random seed 
                                              \\ \cline{2-3} 
                                              & \texttt{max\_depth}                         & {[}1, 2, 3, 5, 7{]}                                                                           \\ \bottomrule
\end{tabular}
} 
\label{tab:xgboost_hyperparams} 
\end{table}

\subsubsection{\texorpdfstring{\textbf{Independent Validation
Cohort}}{Independent Validation Cohort}}\label{independent-validation-cohort}

The study protocol was approved by WCG (Institutional Review Board
\#1371945). This validation cohort is part of a larger, 30-week
longitudinal study designed to assess the impact of lifestyle
modifications on cardiometabolic health metrics. Data collected
throughout the study included anthropometric measurements (e.g., BMI,
waist circumference, skin tone, and blood pressure), blood biomarkers
(such as HbA1c, fasting glucose, fasting insulin, and lipid panel),
wearable device data (Fitbit Charge 6, measuring heart rate, resting
heart rate, heart rate variability, sleep duration, and step count), and
health and lifestyle questionnaires. Participants attended in-person
appointments at the Fitbit Human Research Laboratory in San Francisco at
the beginning and end of the study to complete the aforementioned
measurements, tests, and surveys. All participants were required to wear
a Fitbit Charge 6 device continuously for a minimum of 20 hours per day
throughout the study. A primary hypothesis of this study is that certain
physiological biomarkers, such as insulin resistance, can be predicted
from wearable device data, demographic information, and routine blood
biomarkers.

The inclusion criteria of this study included, participants must be
located in California, at least 18 years of age, able to stand and walk
without aid, willing to provide informed consent, comfortable with
English instructions, own and use a smartphone, willing to download and
use the Fitbit app, comfortable using wearable devices, have regular
internet access, and be willing to comply with study procedures.
Exclusion criteria included, unwillingness to attend lab visits, having
uncontrolled hypertension or recent changes in related medications,
conditions making blood draws impractical, known bloodborne pathogens,
Type 1 Diabetes, Type 2 Diabetes with HbA1c \textgreater{} 8.0\%,
history of gastric surgery, non-iron deficiency anemia, severe mental
illness, heart failure, undergoing dialysis, terminal illness, taking
medications for diabetes or glucose control (with specific examples
provided), taking oral steroids, using tanning lotions on arms, having
wrist tattoos, skin conditions that interfere with devices, pregnancy or
plans for pregnancy, recent radiological procedures with contrast
agents, having non-permissible internal or irremovable external objects,
weight greater than 450 lbs, and unwillingness to participate in test
protocols.

Initially, 144 individuals were enrolled, where 127 individuals had
complete wearable data, and 82 individuals had complete physiological
biomarker data acquired during an in-person visit at the end of the
study. Ultimately, 72 individuals possessed both complete wearable data
and complete physiological biomarker data, and this group was used as an
independent validation cohort for the insulin resistance prediction
model. For the validation we utilized blood biomarkers from the final
study timepoint to ensure the corresponding wearable data reflected a
substantial representation of participants\textquotesingle{} lifestyles.
Wearable data encompassed all available records up to the final visit.
Similarly, demographic data (age and BMI) were extracted from the final
visit.

This final cohort (N=72) was utilized to validate the generalizability
of our insulin resistance prediction models. The independent validation
cohort\textquotesingle s characteristics included an average age of 44.5
years, a BMI of 30.6 kg/m², and a gender distribution of 50 females and
22 males. The cohort comprised individuals of mixed ethnicities: 25
White, 31 Asian, 5 Hispanic or Latino, 4 African American, 6 individuals
who identified as mixed race, and 1 individual who chose not to answer.
Similar to the initial cohort (N=1165), HOMA-IR was used as the ground
truth measure for insulin resistance in the independent validation
study. Using HOMA-IR cut-off values of 2.9 and 1.5, the validation
cohort was divided into three groups: 33 insulin sensitive, 20 with
impaired insulin sensitivity, and 19 with insulin resistance.
\textbf{Table 3} provides detailed characteristics for each insulin
resistance class.

\subsubsection{\texorpdfstring{\textbf{Evaluation of Prediction and
Classification of HOMA-IR}
}{Evaluation of Prediction and Classification of HOMA-IR }}\label{evaluation-of-prediction-and-classification-of-homa-ir}

\textbf{Evaluation of Regression}. The first stage of our framework fh
experiment, we stored the predicted HOMA-IR values in the \emph{test}
set of each fold, computing the average and standard deviation of Mean
Absolute Error (MAE) and Mean Squared Error (MSE) across all test folds.
To compute the over concordance between our predicted values and true
HOMA-IR, we calculated the \(R^{2}\) on all predicted values, which
includes the prediction for all individuals for when they were included
in the test set of each fold.

\textbf{Evaluation of Classification of ``Insulin Resistant'' Group.} To
classify an individual as \emph{insulin resistant}, we use continuous
HOMA-IR values (either predicted or true values for ground truth) and
use the threshold of \(IR = 2.9\) for binary classification. Individuals
with a HOMA-IR value greater than or equal to 2.9 are considered to be
\emph{insulin resistant}, and \emph{not insulin resistant} otherwise. To
evaluate our classification, \emph{Specificity, Sensitivity, Precision,
Area Under Receiver Operating Characteristics Curve (AUROC), and Area
Under Precision-Recall Curve (AUPRC)}. Similar to the regression
evaluation, we report the average and standard deviation of these
metrics across each test fold for each experiment.

\textbf{Time-Dependent Sensitivity and Robustness Analysis.} In order to
determine the robustness of our approach to time-dependant aggregation,
we performed \(n\)-day aggregation of time series on a rolling window
for the entire study duration. The goal of this analysis is to check the
``consistency'' (robustness) of our predictions for various time windows
within the study. We report the consistency of predictions for all
individuals for \(n = \{ 7,\ 14,\ 30,\ 60\}.\)\footnote{Note that we did
  not include 90 and 120 days, since these windows, maximally, would
  result in one or two windows over the study period.} Additionally, we
compute the coefficient of variation of the predicted HOMA-IR values
across the windows and report the results.

\subsubsection{\texorpdfstring{\emph{Insulin Resistance Literacy and
Understanding
Agent}}{Insulin Resistance Literacy and Understanding Agent}}\label{insulin-resistance-literacy-and-understanding-agent-1}

Large Language Models (LLMs) are powerful tools for generating language
which have shown unprecedented potential to improve human-computer
interactions across many domains, including education \citep{Meyer2023-jz},
clinical practice \citep{Singhal2023-ih, mcduff-ddx, Saab2024-mc},
and interpretation of personal health metrics \citep{Cosentino2024-xt, mallinar2025}. Given patient's known challenges of understanding lab results and derived
metrics \citep{Zhang2020-ya}, such
as HOMA-IR, we aimed to develop an LLM-based framework for interpreting
the results, as well as interacting with users for follow-up questions
and recommendations.

Our proposed LLM-based agent is a ReAct (Reasoning and Acting) agent \citep{Yao2022-xq}
capable of multi-step reasoning and planning. ReAct agents represent a
class of LLM-based agents that synergistically combine internal
reasoning with external actions to accomplish complex tasks. Unlike
traditional LLMs that primarily generate text, ReAct agents interleave
textual reasoning traces (``thoughts") with calls to external tools or
information sources (``actions") and reasoning over this multi-step
process (``observe'').

Our IR agent leverages the ReAct iterative process to dynamically plan,
gather necessary information (e.g. from the web), adjust its strategy
based on retrieved facts or computed observations, and ultimately
respond to the incoming health-related query. IR agent's "thoughts"
provide interpretability and allow the agent to decompose complex
problems, while the "actions" provide grounding and interaction with the
real world or specialized knowledge bases, overcoming the limitations of
relying solely on the LLM\textquotesingle s pre-trained knowledge. This
interplay between internal deliberation and external interaction enables
our agent to tackle tasks requiring both reasoning and grounded
knowledge acquisition in real-time. We provide a schematic of our IR
agent in \textbf{Figure 6A}.

\textbf{IR Agent Toolbox}

A key part of our IR Agent is its ability to intelligently select the
set of tools it requires for answering incoming queries. IR agent's
toolbox includes grounding tool (organic web search), arithmetic tools,
a python sandbox as well as our novel HOMA-IR prediction models . These
models and functions serve as a \emph{set of actions} for our model,
ensuring accurate and deterministic computations that reduce the risk of
error and hallucination \cite{Yao2022-xq}. We
describe the set of tools in the Table below.

\begin{table}[]
\caption{Description of the Various Actions Available to the HOMA-IR Literacy and Understanding Agent, Including Machine Learning Models for HOMA-IR Prediction and Utility Functions for Data Processing and Information Retrieval.}
\resizebox{\textwidth}{!}{
\begin{tabular}{lcl}
\hline
Action Name                                                                                             & \multicolumn{1}{l}{Action Type} & Description                                                                                                                               \\ \hline
\begin{tabular}[c]{@{}l@{}}Demographic-Only \\ HOMA-IR Prediction\end{tabular}                          & ML Model                        & \begin{tabular}[c]{@{}l@{}}Predict HOMA-IR value using \\ age and BMI (demographics)\end{tabular}                                         \\ \hline
\begin{tabular}[c]{@{}l@{}}Wearables + Demographics\\ HOMA-IR Prediction\end{tabular}                   & ML Model                        & \begin{tabular}[c]{@{}l@{}}Predict HOMA-IR value using \\ demographics and Wearables features\end{tabular}                                \\ \hline
\begin{tabular}[c]{@{}l@{}}Wearables + Demographics + Fasting Glucose\\ HOMA-IR Prediction\end{tabular} & ML Model                        & \begin{tabular}[c]{@{}l@{}}Predict HOMA-IR value using \\ demographics, Wearables features, and fasting glucose\end{tabular}              \\ \hline
Wearables + Demographics + Lipid Panel + Metabolic Panel HOMA-IR Prediction                             & ML Model                        & \begin{tabular}[c]{@{}l@{}}Predict HOMA-IR value using \\ demographics, Wearables features, lipid panel, and metabolic panel\end{tabular} \\ \hline
Python Sandbox                                                                                          & Execution Environment           & To run simple python code that the agent deems necessary, for example a simple percent change calculator.                                 \\ \hline
Orgnic Web Serch                                                                                        & Function                        & To search the web for and keep track of sources for citation.                                                                             \\ \hline
Comparison Arithmetic Tool                                                                              & Function                        & To compare values and get reletaive difference between them.                                                                              \\ \hline
Plotter/Visualization                                                                                   & Function                        & To visualize HOMA-IR values and contextualize using population statistics.                                                                \\ \hline
\end{tabular}%
} 
\label{tab:my_table} 
\end{table}

\textbf{Implementation of the IR Agent}

We leverage Google DeepMind's open source OneTwo library \citep{onetwo} to
implement the IR agent. OneTwo is a Python framework specifically
designed for facilitating research on prompting strategies for large
foundation models. Crucially, OneTwo\textquotesingle s flexible
execution model supports the complex interaction patterns required by IR
agent, enabling the interleaving of LLM-generated reasoning ("thoughts")
with external "actions" such as web search or executing our train ML
models. OneTwo\textquotesingle s built-in support for tool use and agent
abstractions directly facilitated the creation of our agent that is LLM
agnostic (i.e. can be used other LLMs and not just models developed by
Google).

\subsubsection{\texorpdfstring{\emph{Evaluation of the Insulin
Resistance Literacy and Understanding
Agent}}{Evaluation of the Insulin Resistance Literacy and Understanding Agent}}\label{evaluation-of-the-insulin-resistance-literacy-and-understanding-agent}

To evaluate the IR agent, our goal was to measure the added benefit of
including information about IR (as predicted by our ML models) for
metabolically-relevant queries, when appropriate, compared to LLM
responses that do not include this information. Additionally, we wanted
to assess the accuracy of the IR interpretation and comprehensiveness in
these responses, as evaluated by endocrinologists (``human experts'').

\emph{Human Expert Evaluation Setup}

From our study, we selected five participants with atypical metabolic
profiles, in consultation with an MD who was not part of the evaluation
team, to ask metabolically-relevant questions.

These individuals were selected from the following groups:

\begin{itemize}
\item
  Individuals with clinically-normal HbA1c, but classified as insulin
  resistant
\item
  Obese individuals with sedentary lifestyle (median steps of less than
  8000 and average steps of less than 8000 steps per week) who were
  classified as being insulin resistant
\item
  Obese individuals with an active lifestyle (average step count of
  10000 or more per day, computed on a weekly basis) who were classified
  as insulin resistant.
\item
  Obese individuals with an active lifestyle who were classified as
  insulin sensitive.
\end{itemize}

Using the data from these individuals, we then generated responses using
our IR agent and the same ``base'' LLM at the core of our agent (Gemini
2.0 Flash); note that both models had access to the same data, which
includes demographic, blood biomarkers and wearables features, and were
asked the same questions.

After generating answers to these questions from the IR agent and the
base LLM, we then recruited five practicing endocrinologist to evaluate
each model response "blindly", where the expert evaluators did not know
which response was from the base model or the IR agent, and the design
or the scope of the IR agent was not shared with them to potentially
bias the ratings. More specifically, we asked the endocrinologists to
evaluate the response in two manners:

\textbf{Side-by-side comparison}: This technique is commonly used in the
field to measure the effect of variables (e.g. inclusion or exclusion of
certain features) in generated responses from similar models. The aim
was to measure the benefits provided by our IR agent, including the
addition of IR information, compared with the base LLM without access to
the IR information and the specialized tool. We asked the expert
evaluators to answer the following rubrics while considering responses
side-by-side:

\begin{itemize}
\item
  Q1 {[}Comprehensiveness{]}. Which of the two responses are more
  complete or more comprehensive? (In your opinion, which is a better
  response)
\end{itemize}

\begin{itemize}
\item
  Q2 {[}Trustworthiness{]}. Which response do you find more trustworthy?
\end{itemize}

\begin{itemize}
\item
  Q3 {[}Personalization{]}. Which response better personalizes different
  aspects of health (e.g. lifestyle and cardiovascular health) with
  respect to the query?
\end{itemize}

\textbf{Absolute accuracy of IR Agent}: In addition to side-by-side
comparisons, we aimed to evaluate the accuracy of the IR agent by itself
across different clinically-relevant dimensions \citep{mallinar2025}. Therefore, we asked the
endocrinologists to rate the ``absolute'' following rubric questions:

\begin{itemize}
\item
  Q1 {[}Factuality{]}. Are all the general statements in the response
  (not related to a user\textquotesingle s specific data or condition)
  factually accurate?
\end{itemize}

\begin{itemize}
\item
  Q2 {[}Reference and Interpretation{]}. Does the response reference the
  user\textquotesingle s personal data and interpret it correctly?
\end{itemize}

\begin{itemize}
\item
  Q3 {[}Safety{]}. Is the response free from potentially harmful medical
  advice or recommendations that if acted upon may cause harm to the
  user?
\end{itemize}

\begin{itemize}
\item
  Q4 {[}Grounding{]}. If provided citations, are all citations in the
  response from relevant and verifiable sources?
\end{itemize}

\subsection{\texorpdfstring{Statistical analysis
}{Statistical analysis }}\label{statistical-analysis}

All statistical significance throughout the manuscript have been
determined using Wilcoxon rank-sum test, and multiple testing correction
was performed via Benjamini-Hochberg method.

\subsection{Visualization methods}\label{visualization-methods}

We used Matplotlib and Figma to plot most of the figures.